\documentclass[10pt,twocolumn,letterpaper]{article}

\usepackage[pagenumbers]{cvpr} 
\usepackage[dvipsnames]{xcolor}
\usepackage{pifont}
\usepackage{amssymb}
\usepackage{diagbox}
\usepackage{siunitx}
\usepackage{wrapfig}

\usepackage[symbol]{footmisc}

\definecolor{cvprblue}{rgb}{0.21,0.49,0.74}
\usepackage[pagebackref,breaklinks,colorlinks,citecolor=cvprblue]{hyperref}

\def\paperID{***} 
\def\confName{***}
\def\confYear{2024}

\title{Inverse Neural Rendering for Explainable Multi-Object Tracking}

\author{Julian Ost$^{ 1}$\footnotemark[1]{}   
\qquad
Tanushree Banerjee$^{ 1}$\footnotemark[1]{}  
\qquad
Mario Bijelic$^{1,2}$
\qquad
Felix Heide$^{1,2}$
\\ \\ $^{1}$Princeton University \quad  $^{2}$Torc Robotics
}

\begin{document}


\maketitle
\footnotetext[1]{Indicates equal contribution.}

\begin{abstract}
Today, most methods for image understanding tasks rely on feed-forward neural networks. While this approach has allowed for empirical accuracy, efficiency, and task adaptation via fine-tuning, it also comes with fundamental disadvantages. Existing networks often struggle to generalize across different datasets, even on the same task. By design, these networks ultimately reason about high-dimensional scene features, which are challenging to analyze. This is true especially when attempting to predict 3D information based on 2D images. 
We propose to recast 3D multi-object tracking from RGB cameras as an \emph{Inverse Rendering (IR)} problem, by optimizing via a differentiable rendering pipeline over the latent space of pre-trained 3D object representations and retrieve the latents that best represent object instances in a given input image. 
To this end, we optimize an image loss over generative latent spaces that inherently disentangle shape and appearance properties. We investigate not only an alternate take on tracking but our method also enables examining the generated objects, reasoning about failure situations, and resolving ambiguous cases. We validate the generalization and scaling capabilities of our method by learning the generative prior exclusively from synthetic data and assessing camera-based 3D tracking on the nuScenes and Waymo datasets. Both these datasets are completely unseen to our method and do not require fine-tuning. Videos and code are available \href{ https://light.princeton.edu/inverse-rendering-tracking/}{here}\footnote[2]{\tiny\url{ https://light.princeton.edu/inverse-rendering-tracking/}\label{link}}.
\end{abstract}




\section{Introduction}
The most successful image understanding methods today employ feed-forward neural networks for performing vision tasks, including segmentation~\cite{long2015fully,li2017fully, chen2014semantic}, object detection~\cite{ren2015faster, girshick2015fast, redmon2016yolo, liu2016ssd, ku2018joint, qi2018frustum, zhou2018voxelnet}, object tracking~\cite{sharma2018beyond, kim2021eagermot, zhou2020CenterTrack, chaabane2021deft, yin2021center, weng2020AB3DMOT,pang2021simpletrack} and pose estimation~\cite{wang2019densefusion, xiang2017posecnn}. Typically, these approaches learn network weights using large labeled datasets. At inference time, the trained network layers sequentially process a given 2D image to make a prediction. Despite being a successful approach across disciplines, from robotics to health, and effective in operating at real-time rates, this approach also comes with several limitations: (i) Networks trained on data captured with a specific camera/geography 
\emph{generalize poorly}, (ii) they typically rely on high-dimensional internal feature representations which are \emph{often not interpretable}, making it hard to identify and reason about failure cases, and, (iii) it is challenging to enforce 3D geometrical constraints and priors during inference.

We focus on multi-object tracking as a task that must tackle all these challenges. Accurate multi-object tracking is essential for safe robotic planning. While approaches using LiDAR point clouds (and camera image input) are successful as a result of the explicitly measured depth \cite{pang2021simpletrack,yin2021center,kim2021eagermot,liu2022bevfusion, weng2020gnn3dmot, focalformer3d, bai2021pointdsc}, camera-based approaches to 3D multi-object tracking have only been studied recently~\cite{hu2021QD3DT, wu2021trades, zhou2020CenterTrack, marinello2022triplettrack, chaabane2021deft, nguyen2022multiCamMultiTrack, gladkova2022directtracker, yang2022qtrack,pang2023PFtrack, wang2023StreamPETR}. Monocular tracking methods, typically consisting of independent detection, 3D dynamic models, and matching modules, often struggle as the errors in the distinct modules tend to accumulate. Moreover, wrong poses in the detections can lead to ID switches in the matching process.

We propose an alternative approach that recasts visual inference problems as inverse rendering (IR) tasks, jointly solving them at test time by optimizing over the latent space of a generative object representation. Specifically, we combine object retrieval through the inversion of a rendering pipeline and a learned object model with a 3D object tracking pipeline. This approach allows us to simultaneously reason about an object's 3D shape, appearance, and three-dimensional trajectory from monocular image input only. The location, pose, shape, and appearance parameters corresponding to the anchor objects are then iteratively refined via test-time optimization to minimize the distance between their corresponding generated objects and the given input image. Rather than directly predicting scene and object attributes, we optimize over a latent object representation to synthesize image regions that best explain the observed image. We match the inverse-rendered objects then be matched by comparing their optimized latents. 

Our method hinges on an efficient rendering pipeline and generative object representation at its core. While the approach is not tied to a specific object representation, we adopt GET3D ~\cite{gao2022get3d} as the generative object prior, that \emph{is only trained on synthetic data} to synthesize textured meshes and corresponding images with an efficient differentiable rendering pipeline. Note that popular implicit shape/object representations do either not support class-specific priors~\cite{park2019deepsdf, mildenhall2020nerf}, or require expensive volume sampling~\cite{shen2023gina3d}. 

The proposed method builds on the inductive geometry priors embedded in our rendering forward model, solving \emph{different several tasks simultaneously}. 
Our method refines object pose as a byproduct, merely by learning to represent objects of a given class. Recovering object attributes as a result of inverse rendering also provides \emph{interpretability ``for free''}: once our proposed method detects an object at test time, it can extract the parameters of the corresponding representation alongside the rendered input view. This ability allows for reasoning about failure cases.

We validate that the method naturally exploits 3D geometry priors and \emph{generalizes across unseen domains and unseen datasets}. After training solely on simulated data, we test on nuScenes \cite{caesar2020nuscenes} and Waymo~\cite{sun2020scalability} datasets, and although untrained, we find that our method \emph{outperforms both existing dataset-agnostic multi-object tracking approaches and dataset-specific learned approaches~\cite{zhou2020CenterTrack} when operating on the same detection inputs.}
In summary, we make the following contributions. 
\begin{itemize}
    \item We introduce an inverse rendering method for 3D-grounded monocular multi-object tracking. Instead of formulating tracking as a feed-forward prediction problem, we propose to solve an inverse image fitting problem optimizing over the latent embedding space of generative scene representations.
    \item We analyze the single-shot capabilities and the interpretability of our method using the generated image produced by our method during test-time optimization.
    \item  Trained only on synthetic data, we validate the generalization capabilities of our method by evaluating on unseen automotive datasets, where the method compares favorably to existing methods when provided the same detection inputs.
\end{itemize}

\paragraph{Scope and Limitations} While facilitating inverse rendering, the iterative optimization in our method makes it slower than classical object-tracking methods based on feed-forward networks. We hope to address this limitation in the future by accelerating the forward and backward passes with adaptive level-of-detail rendering techniques.

\section{Related Work}\label{sec:related}
\vspace{0.5\baselineskip}
\noindent
\noindent Object Tracking is a challenging visual inference task that requires the detection and association of multiple objects. Specific challenges include highly dynamic scenes with partial or full occlusions, changes in appearance, and varying illumination conditions~\cite{yilmaz2006object, wu2013online, smeulders2013visual}. In this section, we first review classical tracking methods and deep detection and association methods. Following, we review 3D scene representations and inverse rendering. 

\vspace{0.5\baselineskip}
\noindent
\textbf{3D Object Tracking.}
An extensively investigated line of work proposes tracking by detection, i.e., to solve the task by first detecting scene objects and then learning to find the associations between the detected objects over multiple frames \cite{breitenstein2009robust,kalal2011tracking,bewley2016simple,bergmann2019tracking,wojke2017simple, Wojke2018deep,cao2022observation}. In addition to association, 3D tracking requires the estimation of object pose. Since directly predicting 3D object pose is challenging ~\cite{huang2021joint}, most existing 3D tracking methods rely on some explicit depth measurements in the form of Lidar point clouds~\cite{dewan2016motion,alvarez2019people, yin2021center}, hybrid camera-lidar measurements~\cite{huang2021joint} or stereo information\cite{gladkova2022directtracker,osep2017combined}. Weng \emph{et al.}~\cite{weng2020AB3DMOT} proposed a generic tracking method that combines a 3D Kalman filter and the Hungarian algorithm for matching on an arbitrary object detector. 

Only recent work~\cite{hu2021QD3DT, marinello2022triplettrack, chaabane2021deft,zhou2020CenterTrack,wu2021trades} tackles monocular 3D tracking. Hu \emph{et al.}~\cite{hu2021QD3DT} relies on similarity across different viewpoints to learn rich features for tracking. DEFT~\cite{chaabane2021deft} jointly trains the feature extractor for detection and tracking using the features to match objects between frames. In contrast, Marinello \emph{et al.}~\cite{marinello2022triplettrack} use an off-the-shelf tracker and enhance image features with 3D motion and bounding box information. Zhou \emph{et al.}~\cite{zhou2020CenterTrack} rely on a minimal input of two frames and predicted heatmaps to perform simultaneous detection and tracking. 
Some 3D tracking methods rely on motion models~\cite{scheidegger2018mono, chen2011kalman, nguyen2004fast} such as the Kalman Filter~\cite{kalman1960new}. Recent methods also make use of 
optical flow predictions~\cite{luiten2020track}, learned motion models metrics~\cite{yang2022qtrack}, long short-term memory modules (LSTM)~\cite{hu2021QD3DT, chaabane2021deft, marinello2022triplettrack} and more recently transformer modules~\cite{pang2023PFtrack, wang2023StreamPETR}.
All the above methods rely on a feed-forward image encoder backbone to predict object features. Departing from this approach, we propose a multi-object tracking method that directly optimizes a consistent three-dimensional reconstruction of objects and 3D motion via an inverted graphics pipeline. 

\begin{figure*}[htb!]
    \centering
    \includegraphics[width=0.96\textwidth]{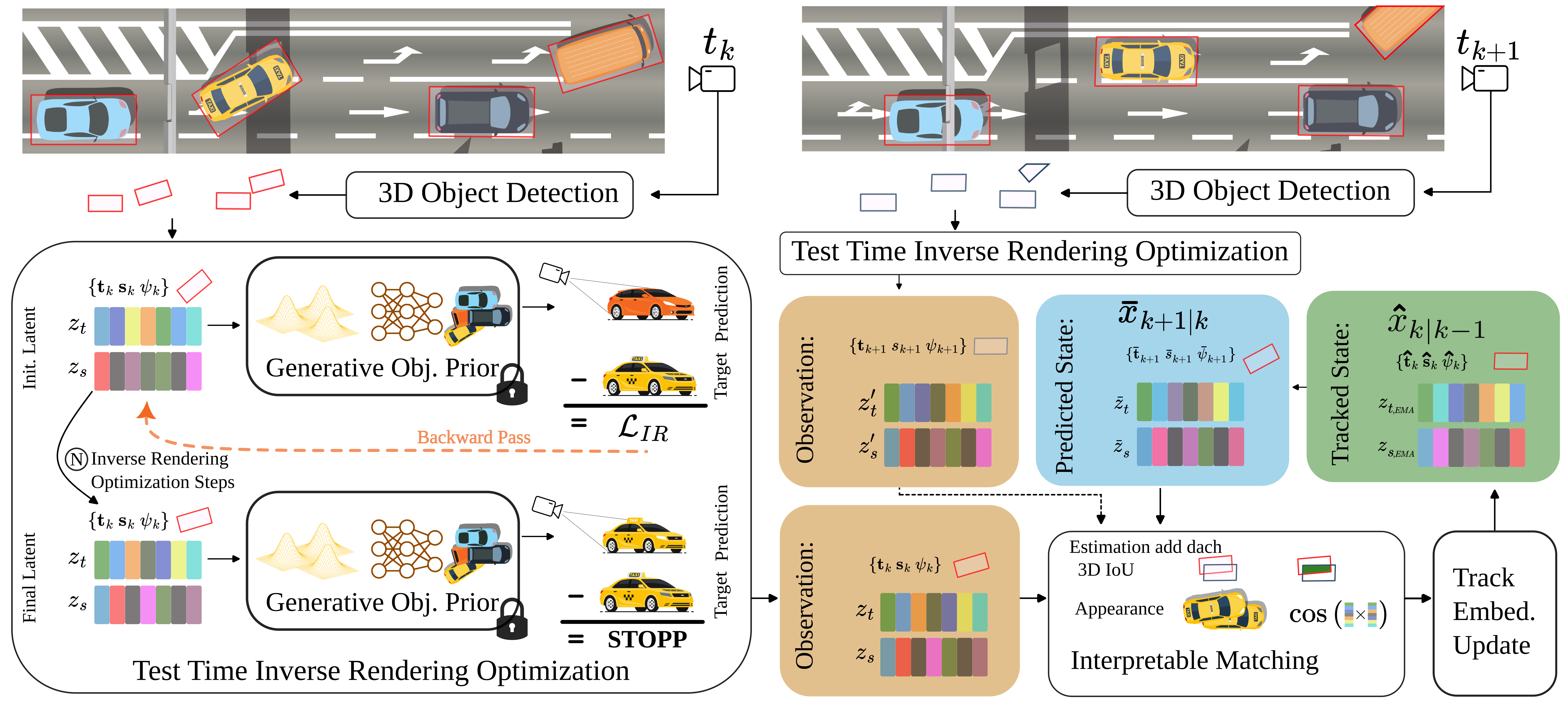}
    \vspace*{-6pt}
    \caption{\textbf{Inverse Rendering for Monocular Multi-Object Tracking.} For each detection, we initialize the embedding codes of an object generator $\mathbf{z}_S$ for shape and $\mathbf{z}_T$ for texture. The generative object prior model is frozen and only embedding codes, pose, and size of each object instance are optimized through inverse rendering to best fit the image observation. Inverse-rendered texture and shape embeddings and refined object locations are provided to the matching stage to match predicted states of tracked objects of the past and new observations. Matched and new tracklets are updated, and unmatched tracklets are ultimately discarded before predicting states in the next step.}
    \label{fig:pipeline}
    \vspace*{-14pt}
\end{figure*}

\vspace{0.5\baselineskip}
\noindent
\textbf{3D Scene Representations, Generation and Neural Rendering.}
A growing body of work addresses joint 3D reconstruction and detection from monocular cameras. Existing methods have exploited different geometrical priors~\cite{mao2022review3dod} for this task, including meshes~\cite{beker2020monocular}, points~\cite{ku2019monocular}, wire frames~\cite{he2019mono3d++}, voxels~\cite{xiang2015voxel3dod} CAD models or implicit functions~\cite{ost2021neural} signed distance functions (SDFs)~\cite{zakharov2020autolabeling}. 
Early approaches in neural rendering represent the scene explicitly by, e.g., encoding texture or radiance on the estimated scene geometry \cite{thies2019deferred} or using volumetric pixels (Voxels) \cite{sitzmann2019deepvoxels}. Other methods represent 3D scenes \emph{implicitly}. This includes the successful NeRF method \cite{mildenhall2020nerf} and variants that have been extended to dynamic scenes~\cite{yuan2021star,ost2021neural,park2021nerfies}.
To allow the handling of semi-transparent objects, these representation models refrain from explicitly representing object surfaces. Signed distance fields represent surfaces of watertight objects as a zero level-set \cite{park2019deepsdf, kellnhofer2021neural, chou2022gensdf} modeling a Signed Distance Function (SDF). 
Adding textures to surface models allows for disentangling object shape from appearance~\cite{xiang2021neutex, koestler2022intrinsic}. In recent years ideas from generative imaging models, such as GANs~\cite{karras2019styleGAN,karras2020styleGAN2}, VAEs and diffusion models~\cite{ho2020denoising,nichol2021improved} have been applied to the 3D domain ~\cite{gao2022get3d,shen2023gina3d,hao2021gancraft,chou2022gensdf}. Generative models can either be used for pure generation~\cite{shen2023gina3d} or provide prior knowledge for downstream tasks. Starting from a good prior can drastically improve the efficiency of inverse tasks, such as IR. While Gina3D~\cite{shen2023gina3d} provides a prior on in-the-wild objects its volumetric rendering pipeline adds another layer of complexity sampling the full volume. We therefore rely on GET3D~\cite{gao2022get3d} generating a mesh as a prior object model and renders through rasterization, profiting from from graphic pipelines optimized over decades. 

\vspace{0.5\baselineskip}
\noindent
\textbf{Inverse Rendering.}
Inverse rendering methods conceptually ``invert'' the graphics rendering pipeline, which generates images from 3D scene descriptions, and instead estimates the 3D scene properties, i.e., geometry, lighting, depth, and object poses based on input images. Recent work~\cite{wang2021nerfmm,yen2021inerf,lin2021barf} successfully achieved joint optimization of a volumetric model and unknown camera poses from a set of images merely by back-propagating through a rendering pipeline. Another area of inverse rendering focuses on material and lighting properties~\cite{nimier2021material,guo2022nerfren, NimierDavidVicini2019Mitsuba2}, to find a representation that best models the observed image.

To the best of our knowledge, we present the first method that employs an inverse rendering approach for multi-object 3D tracking, \emph{without any feed-forward prediction} of object features -- only given 2D image input.
\section{Tracking by Inverse Rendering }\label{sec:method}
We cast object tracking as a test-time inverse rendering problem that fits generated multi-object scenes to the observed image frames. First, we discuss the proposed scene representation we fit. Next, we devise our rendering-based test-time optimization at the heart of the proposed tracking approach. The full tracking pipeline is illustrated in Fig.~\ref{fig:pipeline}. 
We employ an object-centric scene representation. We model the underlying 3D scene for a frame observation as a composition of all object instances without the background scene.

\subsection{Scene Generation}

\noindent \textbf{Object Prior.}  To represent a large, diverse set of instances per class, we define each object instance $o$ as a sample from a distribution $\boldsymbol{O}$ over all objects in a class, that is
\vspace*{-6pt}
\begin{equation}
    \boldsymbol{O} \sim f\left(\boldsymbol{o}\right)\text{,}
\end{equation}
where $f$ is a learned function over a known prior object distribution. Here, the prior distribution is modeled by a differentiable generative 3D object model $o_p = G\left(z_p\right)$, that maps a latent embedding $z_p$ to an object instance $o_p$, the object $p$. In particular, the latent space comprises two disentangled spaces $\boldsymbol{z}_S$ and $\boldsymbol{z}_T$ for shape $S$ and texture $T$.

Given an object-centric camera projection $\mathbf{P}_c = \mathbf{K}_c \mathbf{T}$, where $\mathbf{K}_c $ is the camera intrinsic matrix, a transformation $\mathbf{T} = \left[ \mathbf{R} | \mathbf{t} \right]$ to camera $c$ that is composed of rotation $\mathbf{R}$ and translation $\mathbf{t}$, a differentiable rendering method $R\left(o_p, c\right)$, such as rasterization for meshes or volumetric rendering for neural fields, this renders an image $I_{c,p}$, a 2D observation of the 3D object $o_p$. While our method is general, implementation details of the generator and rendering method are provided in the implementation section.

\vspace{0.5\baselineskip}
\noindent \textbf{Scene Composition.} We model a multi-object scene as a scene graph composed of transformations in the edges and object instances in the leaf nodes, similar to Ost \emph{et al.}~\cite{ost2021neural}. Object poses are described by the homogeneous transformation matrix $\mathbf{T}_{p} \in \mathbb{R}^{4x4}$ with the translation $\mathbf{t}_{p}$ and orientation $\mathbf{R}_{p}$ in the reference coordinate system. The camera pose $\mathbf{T}_{c} \in \mathbb{R}^{4x4}$ is described in the same reference coordinate system. The relative transformation of the camera $c$ and each object instance $p$ can be computed through edge traversal in the scene graph as
\begin{equation}\label{eq:camera_to_obj}
    \mathbf{T}_{c,p} = \text{diag}\left(\frac{1}{s_{p}}\right)\mathbf{T}_{p}\mathbf{T}_{c}^{-1},
\end{equation}
where the factor $s_{p}$ is a scaling factor along all axes to allow a shared object representation of a unified scale. This canonical object scale is necessary to represent objects of various sizes, independent of the learned prior on shape and texture. The object centric projection $\mathbf{P}_{c,p} = \mathbf{K}_c \mathbf{T}_{c,p} $ is used to render the RGB image $I_{c,p} \in \mathcal{R}^{H \times W \times 3}$ and mask $M_{c, p} \in \left[0, 1 \right]^{H \times W}$ for each object/camera pair. 

Individual rendered RGB images are ordered by object distance $\| \mathbf{t}_{c, p} \|$, such that $p=1$ is the shortest distance to $c$. Using the Hadamard product of the non-occluded mask $\gamma_p$ all $N_o $, object images are composed into a single image
\begin{equation}\label{eq:forward_scene}
    \begin{aligned}
            \hat I_c &= \sum_{k = 1}^{N_o} R\left( G\left( \mathbf{z}_{S,p}, \mathbf{z}_{T,p} \right), \mathbf{P}_{c,p}\right) \circ \mathbf{\gamma}_p, \text{\,\,where  } \\        
             \mathbf{\gamma}_p  & = \max \left( \left( \mathbf{M}_{c, p} - \sum_{q=1}^{p} \mathbf{M}_{c,q} \right), \mathbf{0}^{H\times W} \right),
    \end{aligned}
\end{equation}
where instance masks are generated in the same fashion. 

\subsection{Inverse Multi-Object Scene Rendering.}\label{ssec:scene_rep}
%
We invert the differentiable rendering model defined in Eq.~\ref{eq:forward_scene} by optimizing the set of all object representations in a given image $I_c$ with gradient-based optimization. We assume that, initially, each object $o_p$ is placed at a pose $\hat{\mathbf{T}}_{c,p}$ and scaled with $\hat s_p$ near its underlying location. We represent object orientations in their respective Lie algebraic form $\mathfrak{so}(3)$. We further sample an object embedding $\hat{\mathbf{z}}_{S,p}$ and $\hat{\mathbf{z}}_{T,p}$ in the respective latent embedding space.

For in-the-wild images, $I_c$ is not just composed of sampled object instances but other objects and the scene background. Since our goal for tracking is the reconstruction of all object instances of specific object classes, a na\"ive $\ell_2$ image matching objective of the form $\| I_c - \hat{I}_c \|_2$ is noisy and challenging to solve with vanilla stochastic gradient descent methods. To tackle this issue, we optimize visual similarity in the generated object regions instead of the full image. We optimize only on rendered RGB pixels and minimize
\begin{equation}\label{eq:loss_mse}
    \begin{aligned}
        \mathcal{L}_{RGB} = \| \left( I_c - \hat{I}_c \right) \circ \hat{M}_{I_c} \|_2, \\ \text{\,\,with\ } \hat{M}_{I_c} = \min \left( \sum M_{c,p} , \mathbf{1}\right).
    \end{aligned}
\end{equation}
The mask of all foreground/object pixels $\hat{M}_{I_c}$ is computed as the sum over all object masks $M_{c,p}$ in the frame rendered by camera $c$.
We employ a learned perceptual similarity metric~\cite{zhang2018perceptual} (LPIPS) on object-centered image patches, that is
\begin{equation}\label{eq:loss_lpips}
    \mathcal{L}_{perceptual} = \text{LPIPS}_{patch}\left( I_c, \hat{I}_{c,p}  \right).
\end{equation}
The combined loss function of our method is
\begin{equation}\label{eq:loss_mse_lpips}
    \mathcal{L}_{IR} = {L}_{RGB} + \lambda \mathcal{L}_{perceptual},
\end{equation}
which we optimize by refining the latent codes of shape and appearance, position, rotation, and scale, leading to
\begin{equation}
        \hat{\mathbf{z}}_{S,p}, \hat{\mathbf{z}}_{T,p}, \hat{s}_{p} \hat{\mathbf{t}}_p, \hat{\mathbf{R}}_p = \text{arg min} \left( \mathcal{L}_{IR} \right) .
\end{equation}
Instead of using vanilla stochastic gradient descent methods, we propose an alternating optimization schedule of distinct properties that includes aligning $z_S$ before $z_T$, to reduce the number of total optimization steps. A detailed implementation and validation of all design choices of the optimization are presented in the Supplementary Material.

\begin{figure*}[t!]
	\centering
	\resizebox{1.\linewidth}{!}{
	\renewcommand{\arraystretch}{0.5}
	\begin{tabular}{@{}c@{\hskip 0.05cm}c@{\hskip 0.05cm}c@{\hskip 0.05cm}c@{\hskip 0.05cm}c@{\hskip 0.05cm}c@{}}
		\centering
            &
		{\small Input $t_0$}&
		{\small Tracked $t_0$}&
		{\small Tracked $t_1$}&
		{\small Tracked $t_2$}&
		{\small Tracked $t_3$}\\

            \rotatebox[origin=c]{90}{{\footnotesize	  Suburban}}&
		 \raisebox{-0.5\height}{\includegraphics[width=.32\columnwidth, trim={0cm 0cm 0cm 0cm},clip]{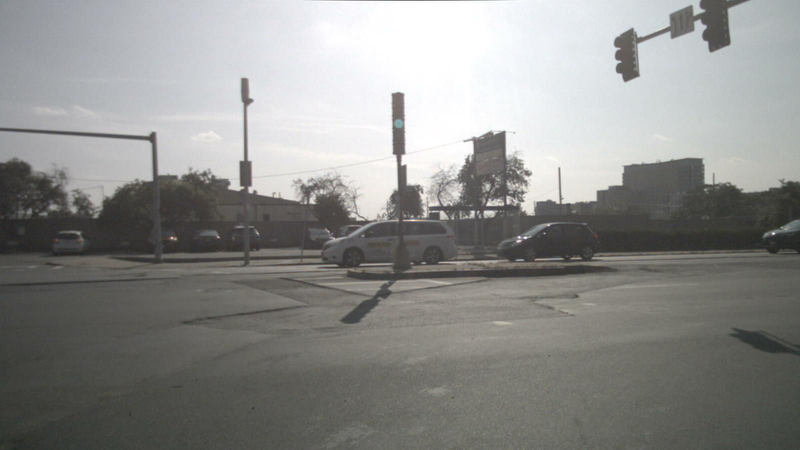}}&
		 \raisebox{-0.5\height}{\includegraphics[width=.32\columnwidth, trim={0cm 0cm 0cm 0cm},clip]{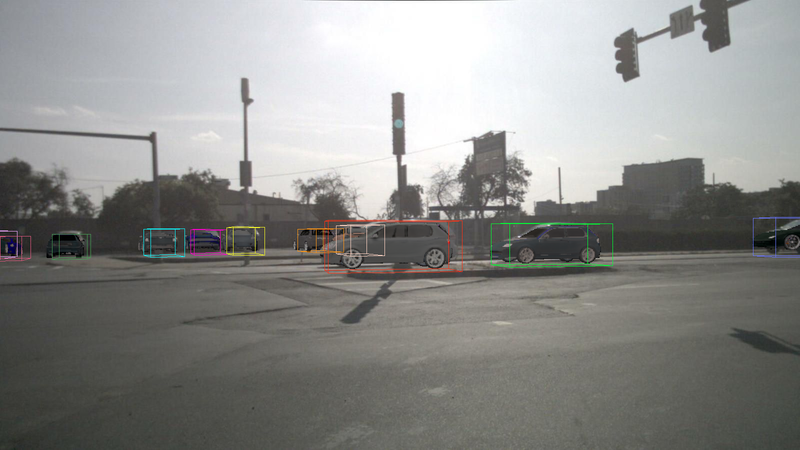}}&
		 \raisebox{-0.5\height}{\includegraphics[width=.32\columnwidth, trim={0cm 0cm 0cm 0cm},clip]{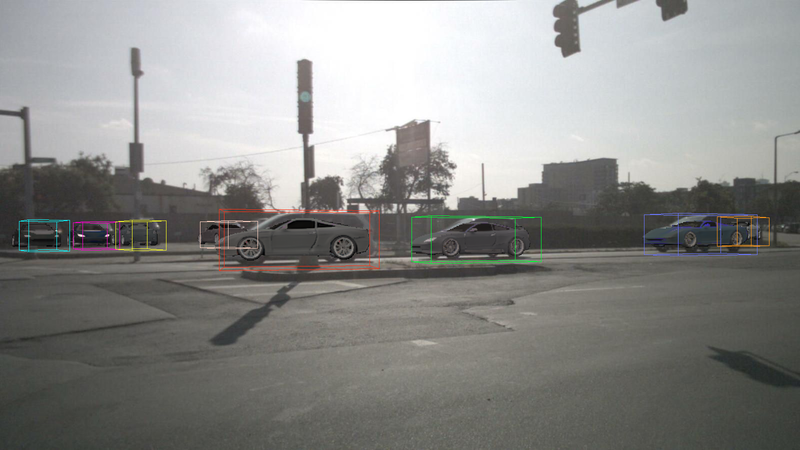}}&
		 \raisebox{-0.5\height}{\includegraphics[width=.32\columnwidth, trim={0cm 0cm 0cm 0cm},clip]{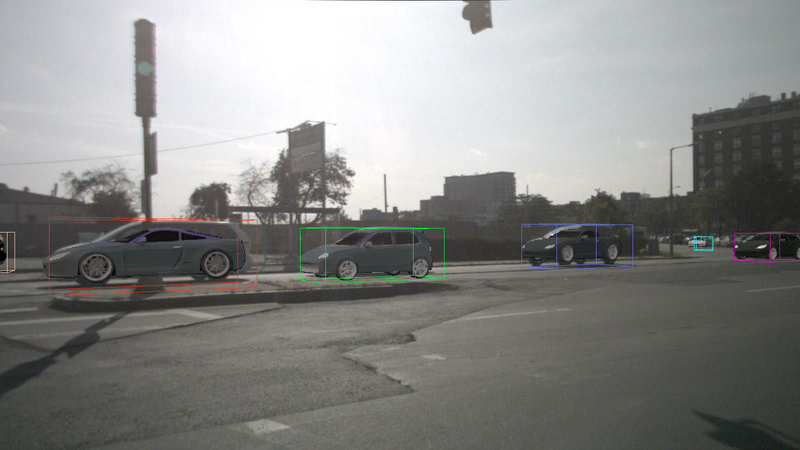}}&
		 \raisebox{-0.5\height}{\includegraphics[width=.32\columnwidth, trim={0cm 0cm 0cm 0cm},clip]{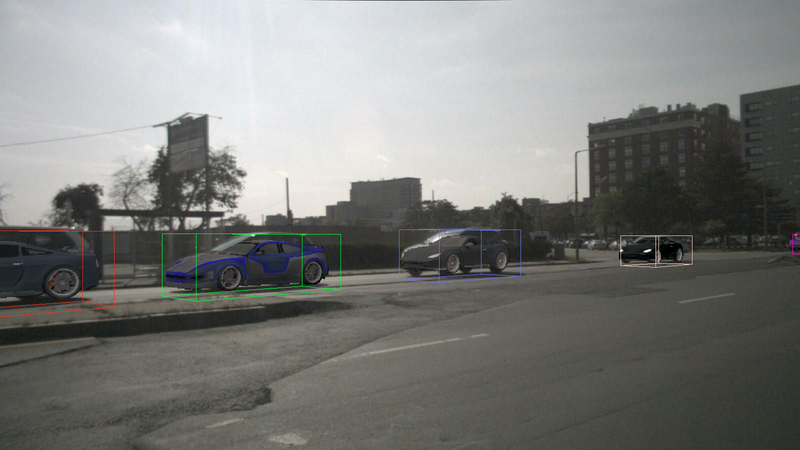}}\\[0.75cm]
  
		\rotatebox[origin=c]{90}{{\footnotesize	 Urban}}&
		\raisebox{-0.5\height}{\includegraphics[width=.32\columnwidth, trim={0cm 0cm 0cm 0cm},clip]{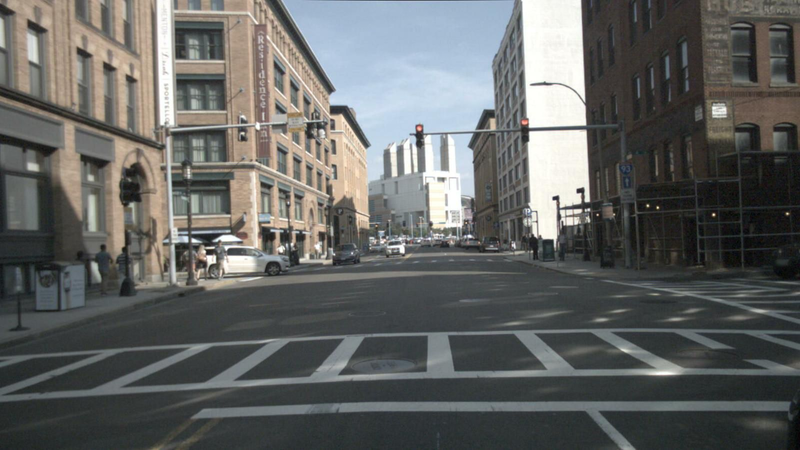}}&
		\raisebox{-0.5\height}{\includegraphics[width=.32\columnwidth, trim={0cm 0cm 0cm 0cm},clip]{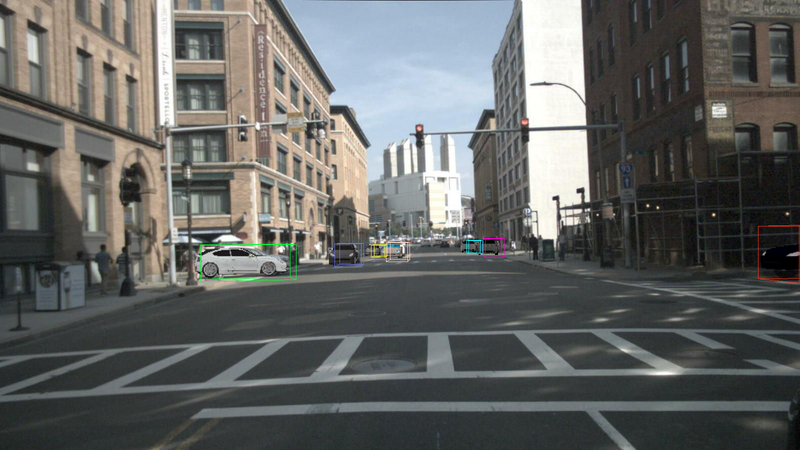}} &
		\raisebox{-0.5\height}{\includegraphics[width=.32\columnwidth, trim={0cm 0cm 0cm 0cm},clip]{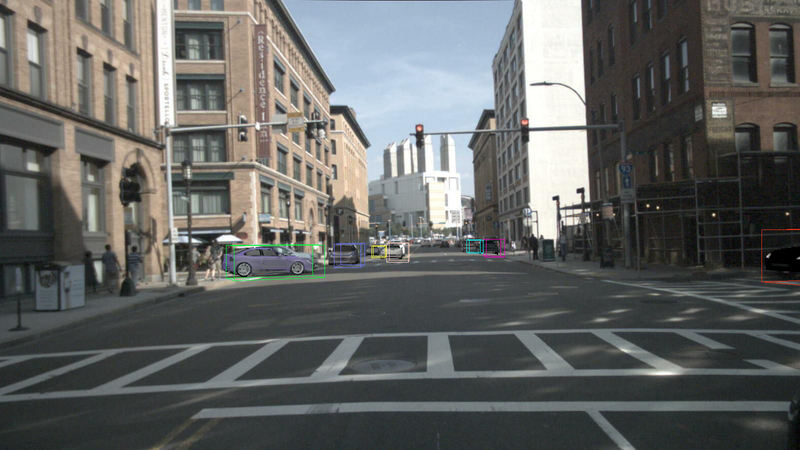}}&
		\raisebox{-0.5\height}{\includegraphics[width=.32\columnwidth, trim={0cm 0cm 0cm 0cm},clip]{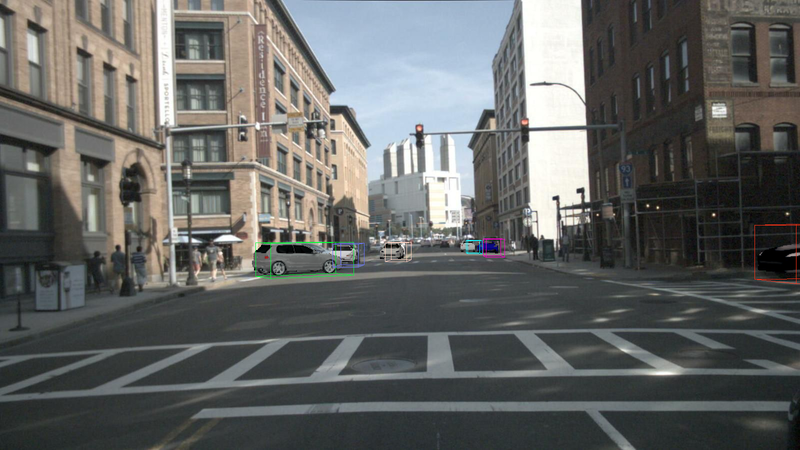}}&
		\raisebox{-0.5\height}{\includegraphics[width=.32\columnwidth, trim={0cm 0cm 0cm 0cm},clip]{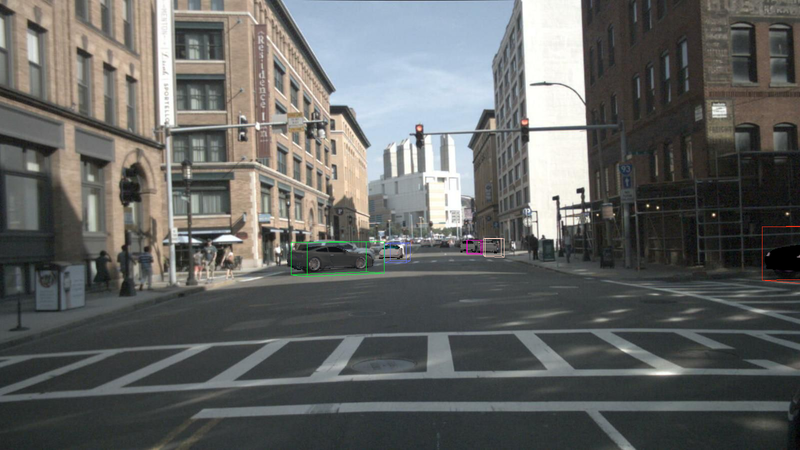}}\\[0.75cm]
  
            \rotatebox[origin=c]{90}{{\footnotesize	 Clutter}}&
  		\raisebox{-0.5\height}{\includegraphics[width=.32\columnwidth, trim={0cm 0cm 0cm 0cm},clip]{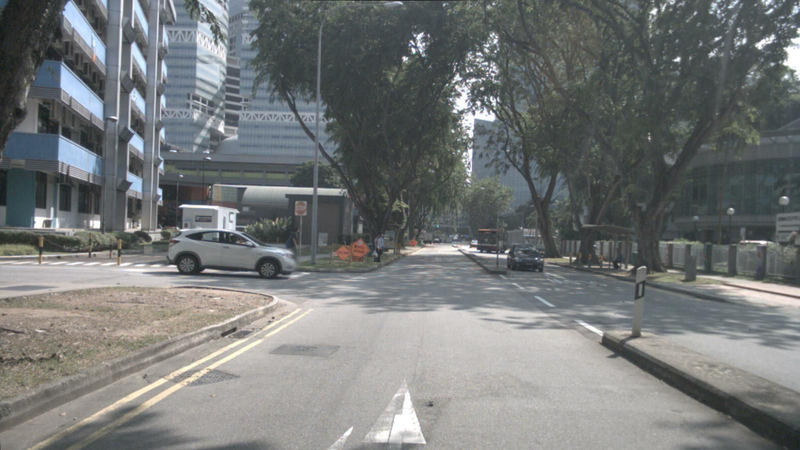}}&
		\raisebox{-0.5\height}{\includegraphics[width=.32\columnwidth, trim={0cm 0cm 0cm 0cm},clip]{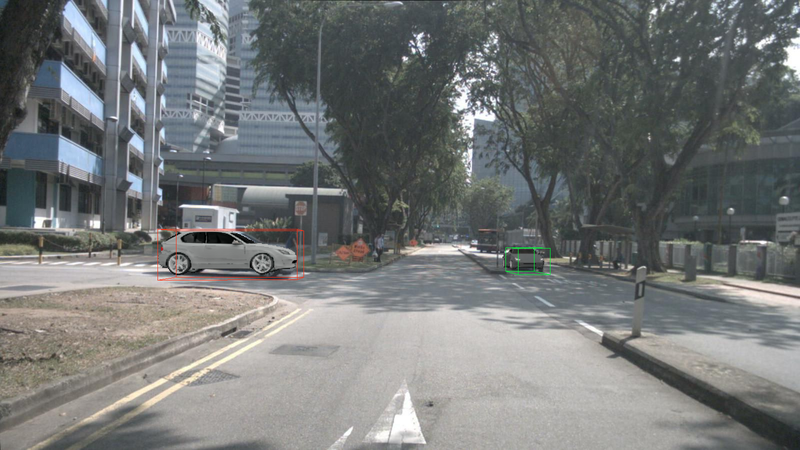}}&
		\raisebox{-0.5\height}{\includegraphics[width=.32\columnwidth, trim={0cm 0cm 0cm 0cm},clip]{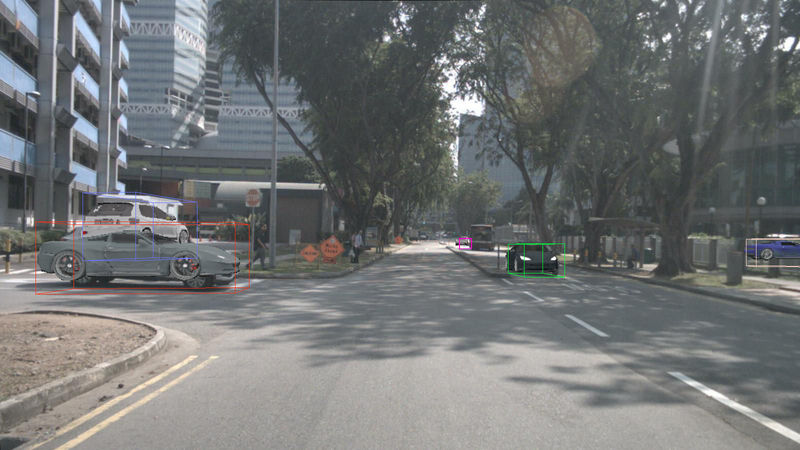}}&
		\raisebox{-0.5\height}{\includegraphics[width=.32\columnwidth, trim={0cm 0cm 0cm 0cm},clip]{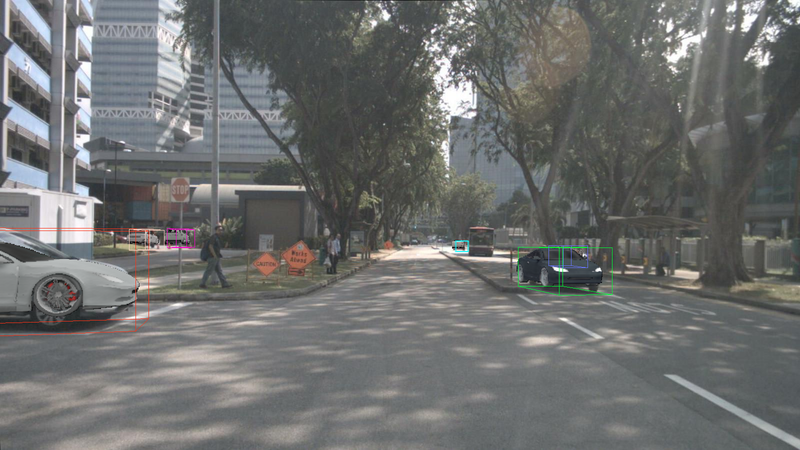}}&
		\raisebox{-0.5\height}{\includegraphics[width=.32\columnwidth, trim={0cm 0cm 0cm 0cm},clip]{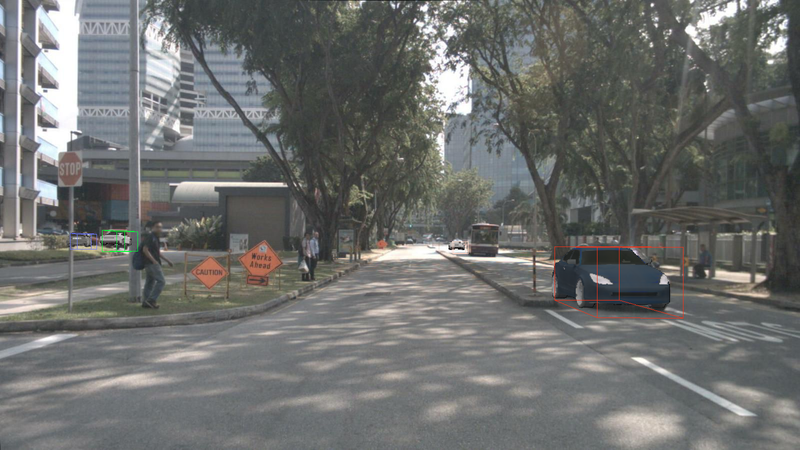}}\\[0.75cm]
  
            \rotatebox[origin=c]{90}{{\footnotesize	Parking lot}}&
  		\raisebox{-0.5\height}{\includegraphics[width=.32\columnwidth, trim={0cm 0cm 0cm 0cm},clip]{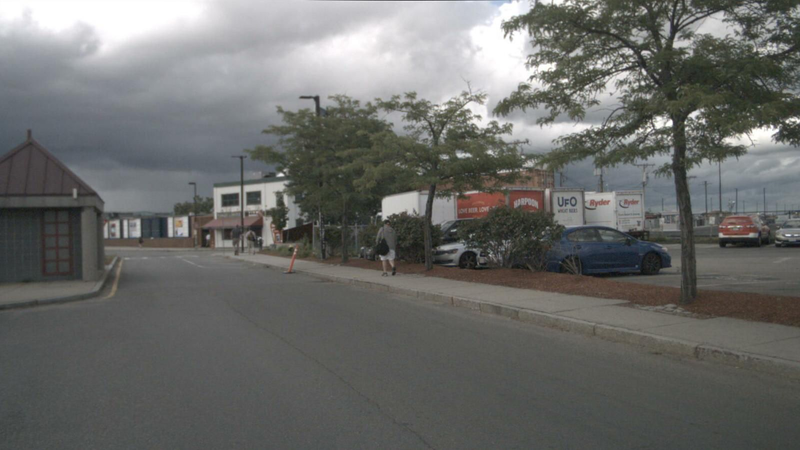}}&
		\raisebox{-0.5\height}{\includegraphics[width=.32\columnwidth, trim={0cm 0cm 0cm 0cm},clip]{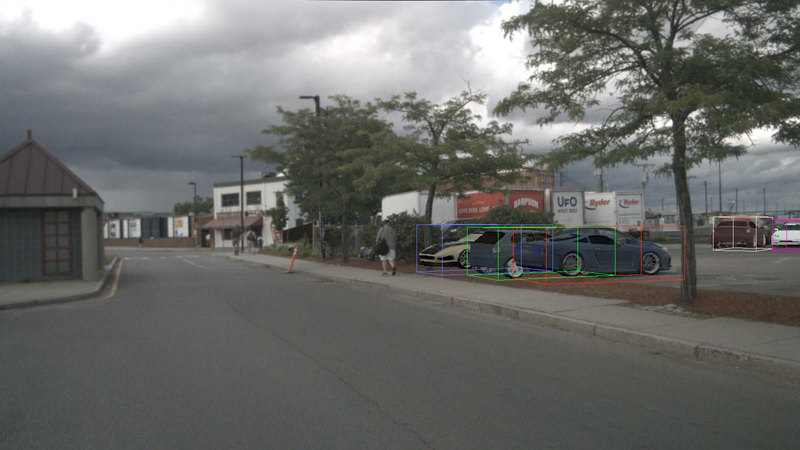}}&
		\raisebox{-0.5\height}{\includegraphics[width=.32\columnwidth, trim={0cm 0cm 0cm 0cm},clip]{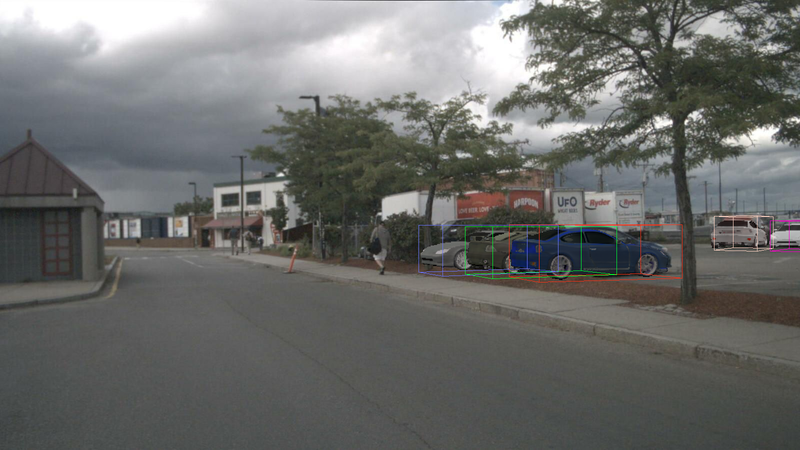}}&
		\raisebox{-0.5\height}{\includegraphics[width=.32\columnwidth, trim={0cm 0cm 0cm 0cm},clip]{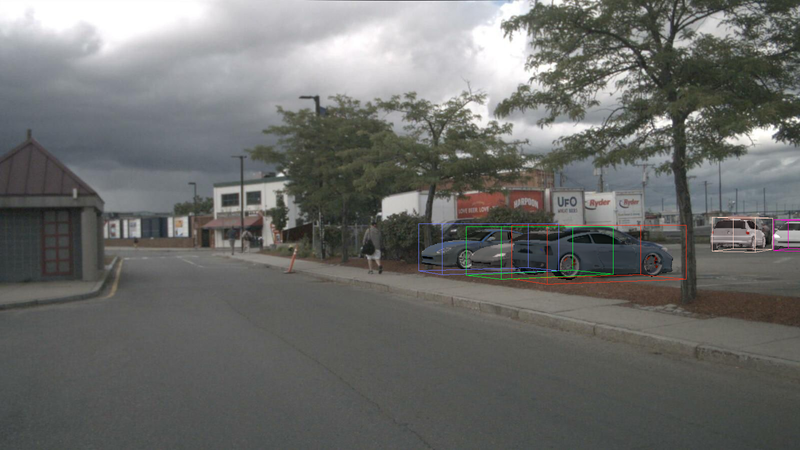}}&
		\raisebox{-0.5\height}{\includegraphics[width=.32\columnwidth, trim={0cm 0cm 0cm 0cm},clip]{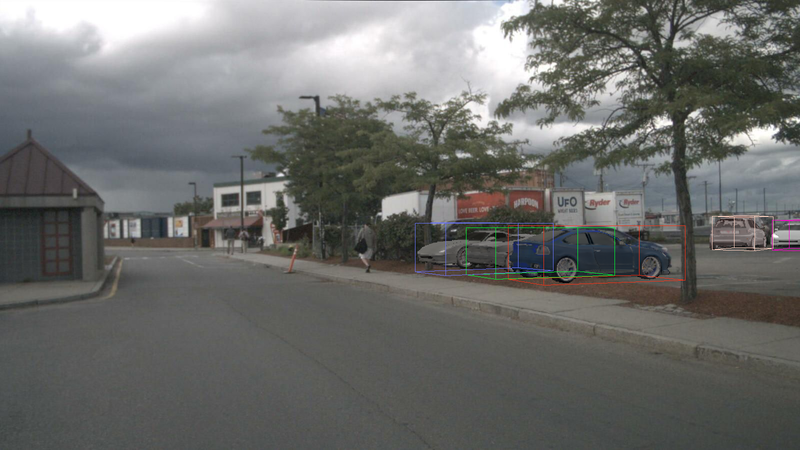}}\\
		
	\end{tabular}
	}
\vspace*{-6pt}
\caption{Tracking via Inverse Neural Rendering on nuScenes~\cite{caesar2020nuscenes}. From left to right, we show (i) observed images from diverse scenes at timestep $k=0$; (ii) an overlay of the optimized generated object and its 3D bounding boxes at timestep $k=0, 1, 2 \text{ and } 3$. The color of the bounding boxes for each object corresponds to the predicted tracklet ID. We see that even in such diverse scenarios, our method does not lose any tracks and performs robustly across all scenarios, although the dataset is unseen.}\label{fig:nuScenes_results}
\vspace*{-12pt}
\end{figure*}
\vspace{0.5\baselineskip} \noindent \textbf{Optimization.}
To solve Eq.~\ref{eq:loss_mse_lpips}, we propose an optimization schedule, that first optimizes a coarse color, and then jointly optimizes the shape and the positional state of each object. As the backbone of the learned perceptual loss, we use a pre-trained VGG16~\cite{simonyan2015deep} and utilize individual output feature map similarities at different points of the optimization. We find that color and other low-dimensional features are represented in the initial feature maps and those are better guidance for texture than high-dimensional features as outputs of the later blocks. These features have a more informative signal for shape and object pose. We use the average of the first and second blocks in the optimization for $\mathbf{z}_T$, while the combined perceptual similarity loss guides the optimization of  $\mathbf{z}_T$ and the pose.

We initialize all object embeddings with the same fixed values inside the embedding space, take two optimization steps solely on color utilizing the described loss, and then freeze the color for the joint optimization of the shape and pose. We regularize out-of-distribution generations with
\begin{equation}\label{eq:regularize}
    \mathcal{L}_{embed} = \alpha_T\mathbf{z}_T + (1 - \alpha_T)\mathbf{z}_T^{avg} +  \alpha_S\mathbf{z}_S + (1 - \alpha_S)\mathbf{z}_S^{avg}
\end{equation}
that minimizes a weighted distance in each dimension with respect to the average embedding  $\mathbf{z}_S$ or $\mathbf{z}_T$ respectively. For optimization, we use the ADAM optimizer \cite{Kingma2015AdamAM}.
The final loss function combines the RGB, perceptual cost  and the regularization with $\lambda=10$, $\alpha_{T}=0.7$ and $\alpha_{S}=0.7$ of Eq.~\ref{eq:loss_mse_lpips} and Eq.~\ref{eq:regularize}. We freeze color after two steps of optimization and optimize the shape and scale for three more steps, adding translation and rotation only in the last two steps.

\subsection{3D MOT via Inverse Rendering}
Next, we describe the proposed method for tracking multiple dynamic objects with the inverse rendering approach from above. The approach tracks objects in the proposed representation across video frames and is illustrated in Fig.~\ref{fig:pipeline}. For readability, we omit $p$ and the split of $\mathbf{z}$ into $\mathbf{z}_S$ and $\mathbf{z}_T$ in the following.

\vspace{0.5\baselineskip} \noindent \textbf{Initial Object and Pose Estimation.}
Common to tracking methods, we initialize with a given initial 3D detection on image $I_{c,k}$, and we set object location $\mathbf{t}_{k} = \left[x, y, z\right]_{k}$, scale $s_{k} = \max(\text{w}_{k}, \text{h}_{k}, \text{l}_{k})$ using the detected bounding box dimensions and heading $\psi_{k}$ in frame $k$. We then find an optimal representation $\mathbf{z}_k$, and a refined location and rotation of each object $o$ via the previously introduced inverse rendering pipeline for multi-object scenes. The resulting location, rotation, and scale lead to the observation vector 
\begin{equation}\label{eq:observation}
\mathbf{y}_{k} = [\mathbf{t}_{k}, s_{k},\psi_{k}] .
\end{equation}

\vspace{0.5\baselineskip} \noindent \textbf{Prediction.}
While not confined to a specific dynamics model, we use a linear state-transition model $\textbf{A}$, for the objects state $\mathbf{x}_{k} = [x, y, z, s, \psi, w, h, l, {x}', {y}', {z}']_{k}$, and a forward prediction using a Kalman Filter~\cite{kalman1960new}, a vanilla approach in 3D object tracking~\cite{weng2020AB3DMOT}. An instantiated object in k-1 can be predicted in frame $k$ as
\begin{equation}
\begin{aligned}
   &\hat{\mathbf{x}}_{k|k-1} = \mathbf{A}\hat{\mathbf{x}}_{k-1|k-1} \\ &\text{ and } \mathbf{P}_{k|k-1} = \mathbf{A}\mathbf{P}_{k-1|k-1}\mathbf{A}^T + \mathbf{Q}, 
\end{aligned}
\end{equation}
with the predicted \textit{a priori} covariance matrix modeling the uncertainty in the predicted state.

\begin{figure*}[t!]
	\centering
\resizebox{1.\linewidth}{!}{
	\renewcommand{\arraystretch}{0.5}
	\begin{tabular}{@{}c@{\hskip .05cm}c@{\hskip .05cm}c@{\hskip .05cm}c@{\hskip .05cm}c@{\hskip .05cm}c@{}}
		\centering
            &
		{\small Input $t_0$}&
		{\small Tracked $t_0$}&
		{\small Tracked $t_1$}&
		{\small Tracked $t_2$}&
		{\small Tracked $t_3$}\\

            \rotatebox[origin=c]{90}{{\footnotesize	Highway}}&
		\raisebox{-0.5\height}{\includegraphics[width=.32\columnwidth, trim={0cm 0cm 0cm 0cm},clip]{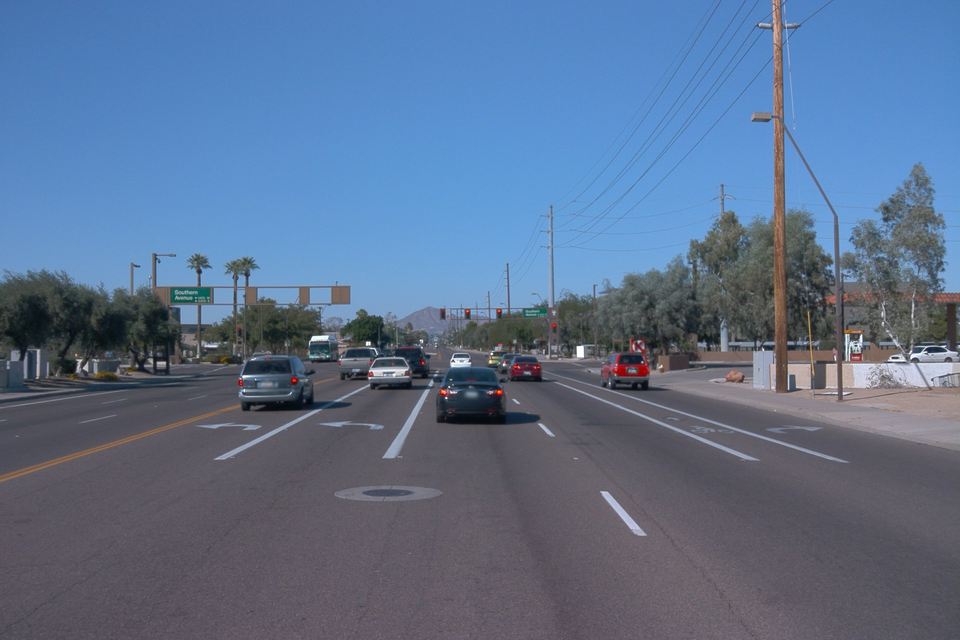}}&
		\raisebox{-0.5\height}{\includegraphics[width=.32\columnwidth, trim={0cm 0cm 0cm 0cm},clip]{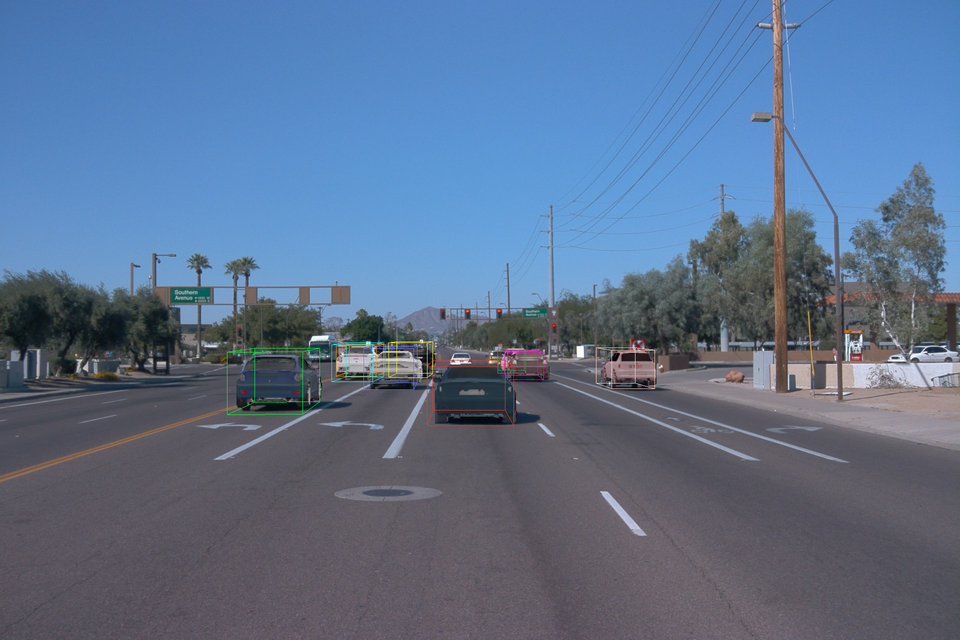}}&
		\raisebox{-0.5\height}{\includegraphics[width=.32\columnwidth, trim={0cm 0cm 0cm 0cm},clip]{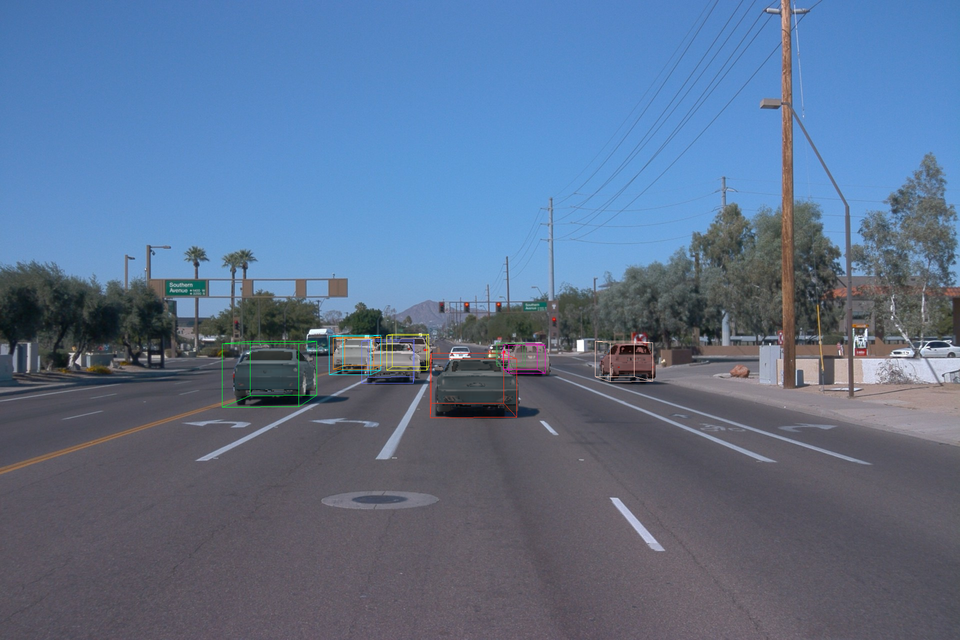}}&
		\raisebox{-0.5\height}{\includegraphics[width=.32\columnwidth, trim={0cm 0cm 0cm 0cm},clip]{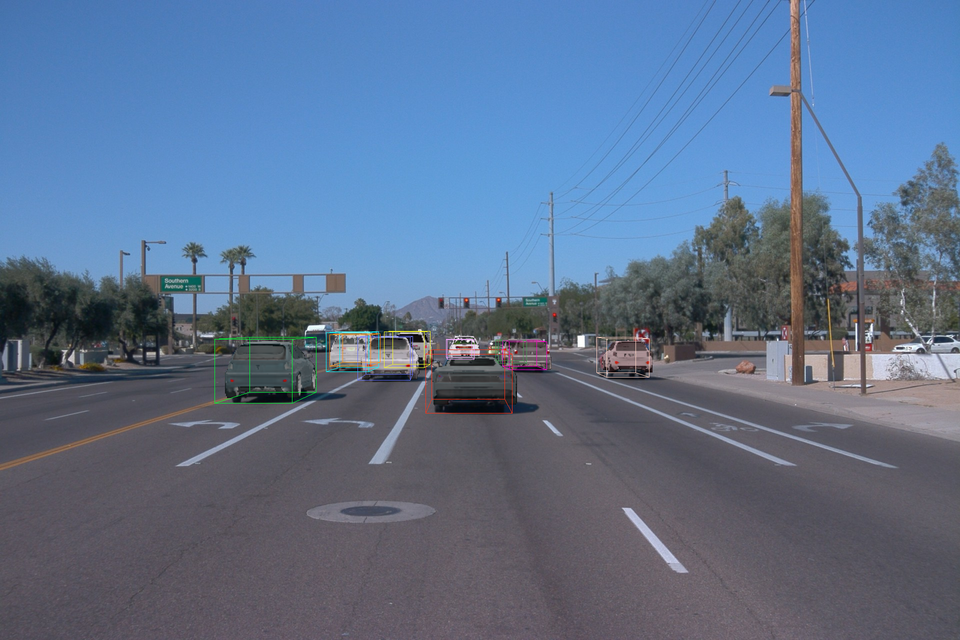}}&
		\raisebox{-0.5\height}{\includegraphics[width=.32\columnwidth, trim={0cm 0cm 0cm 0cm},clip]{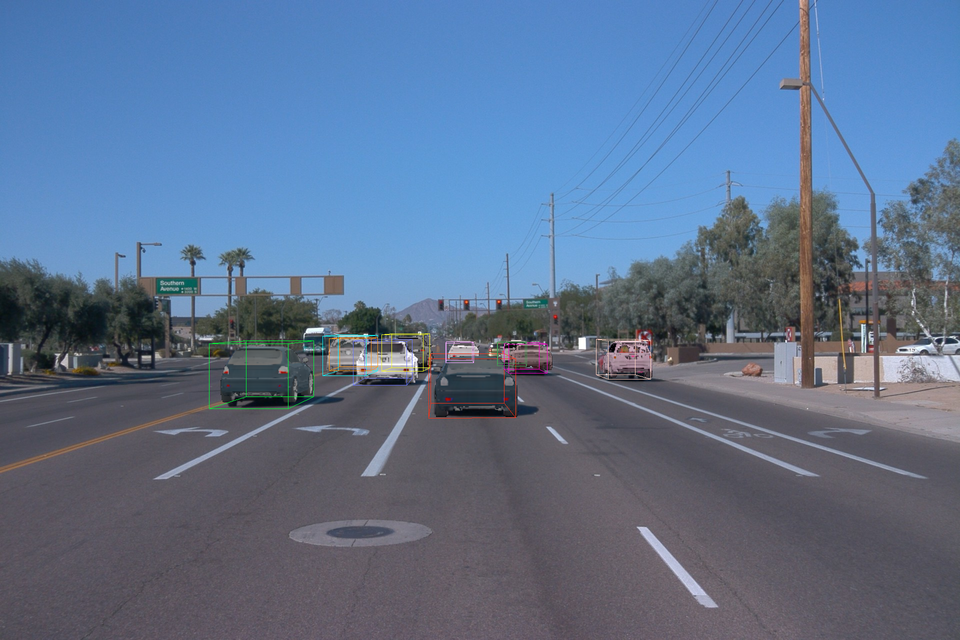}}\\[0.9cm]

            \rotatebox[origin=c]{90}{{\footnotesize	Urban}}&
		\raisebox{-0.5\height}{\includegraphics[width=.32\columnwidth, trim={0cm 0cm 0cm 0cm},clip]{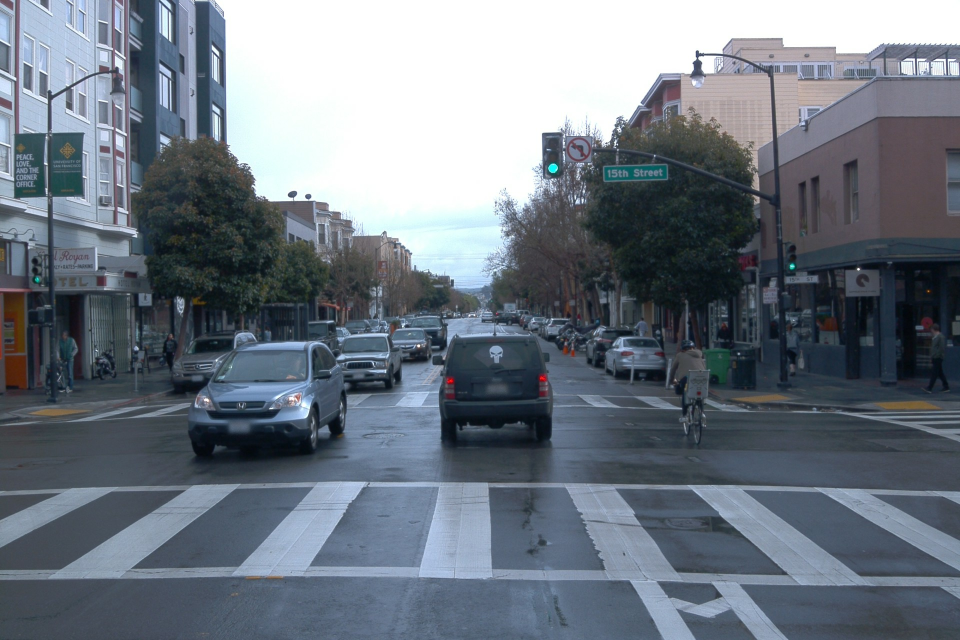}}&
		\raisebox{-0.5\height}{\includegraphics[width=.32\columnwidth, trim={0cm 0cm 0cm 0cm},clip]{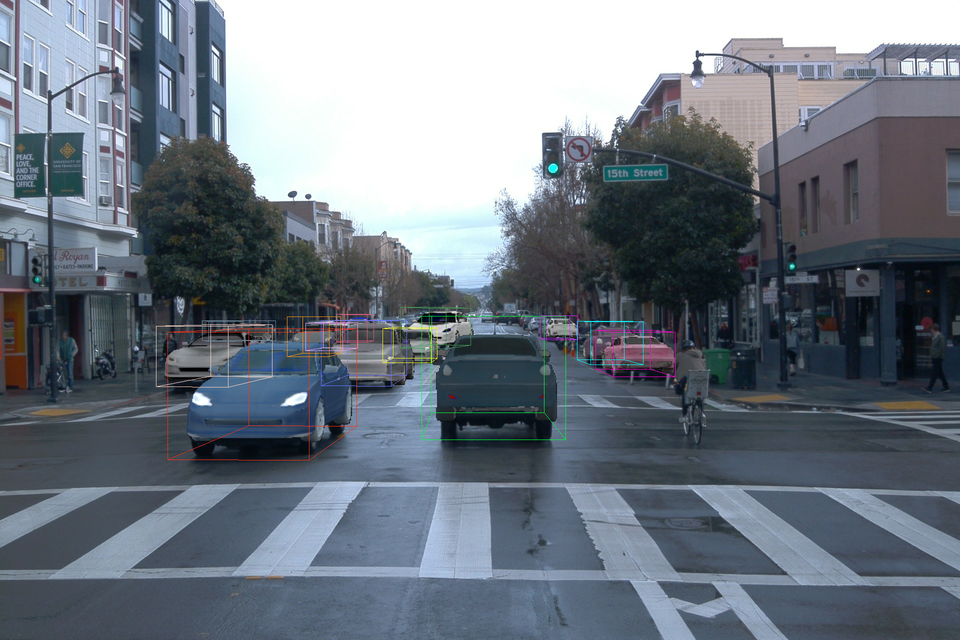}}&
		\raisebox{-0.5\height}{\includegraphics[width=.32\columnwidth, trim={0cm 0cm 0cm 0cm},clip]{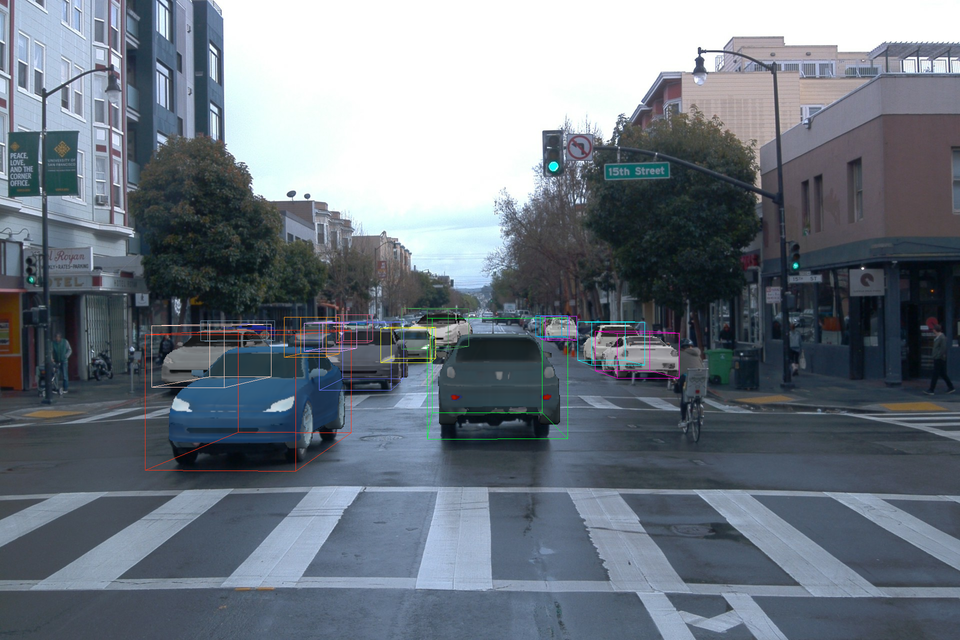}}&
		\raisebox{-0.5\height}{\includegraphics[width=.32\columnwidth, trim={0cm 0cm 0cm 0cm},clip]{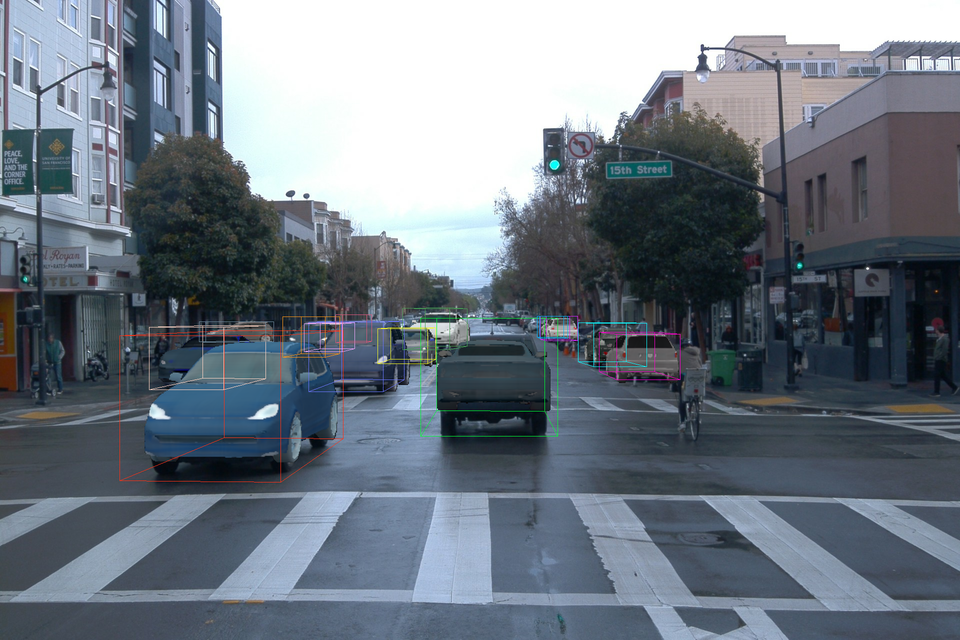}}&
		\raisebox{-0.5\height}{\includegraphics[width=.32\columnwidth, trim={0cm 0cm 0cm 0cm},clip]{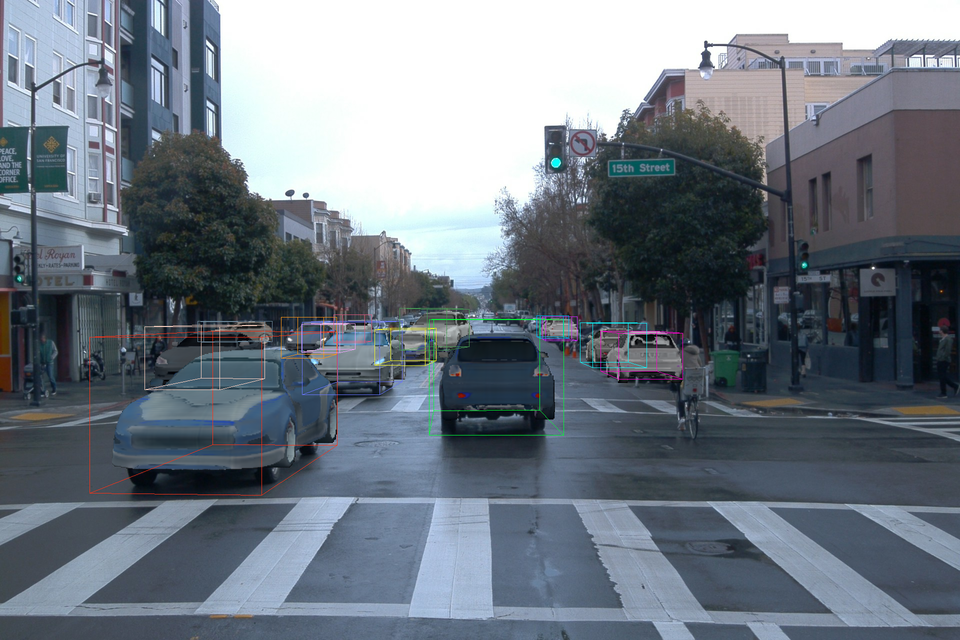}}\\

	\end{tabular}
	}
\vspace{-6pt}
\caption{Without changing the model or training on the dataset, our proposed method can generalize well to the Waymo Open Driving Dataset ~\cite{sun2020scalability}. Similar to Fig \ref{fig:nuScenes_results}, from left to right, we show (i) observed images from diverse scenes from the dataset at timestep $k=0$; (ii) an overlay of the closest generated object and predicted 3D bounding boxes at timestep $k=0, 1, 2 \text{ and } 3$. The color of the bounding boxes for each object corresponds to the predicted tracklet ID. Our method does not lose any tracks even on a different unseen dataset in diverse scenes, validating that the approach generalizes.}\label{fig:waymo_results}
\end{figure*}

\vspace{0.5\baselineskip} \noindent \textbf{Interpretable Latent Matching.}
%
In the matching stage, all optimal object representations $o_p$ in frame $k$ are matched with \emph{tracked} and \emph{lost} objects from $k-1$. Objects are matched based on appearance and location with a weighted affinity score
\begin{equation}\label{eq:affinity}
    A = w_{IoU} A_{IoU,3D} + w_{z} A_{z} + w_{c} D_{centroid},
\end{equation}
where $A_{IoU,3D}$ is the IoU of the 3D boxes computed over the predictions of tracked object predictions $\mathbf{x}_{k|k-1}$ and refined observations. Here, the object affinity $A_{z}$ is computed as the cosine distance of tracked object latent embeddings $\mathbf{z}$. In addition to that the Euclidean distance between the center $D_{centroid}$ adds additional guidance. We add no score for unreasonable distant tracked objects and detections.

We compute the best combination of tracked and detected objects using the Hungarian algorithm~\cite{kuhn1955hungarian}, again a conventional choice in existing tracking algorithms. Matched tracklet and object pairs are kept in the set of \emph{tracked} objects and the representation of the corresponding detections is discarded, while unmatched detections are added as new objects.
Unmatched tracklets are set to \emph{lost} with a lost frame counter of one. Objects that were not detected in previous frames are set to \emph{tracked} and their counter is reset to $0$. Objects with a lost frame count higher than lifespan $N_{life}$, or outside of the visible field, are removed.

\vspace{0.5\baselineskip} \noindent \textbf{Track and Embedding Update.}
In the update step, we refine each object embedding $\mathbf{z}$  and motion model $\mathbf{y}_{k}$ given the result of the matching step. Embeddings are updated through an exponential moving average
\vspace*{-4pt}
\begin{equation}
    \mathbf{z}_{k, EMA}  = \beta \mathbf{z}_k + (1 - \beta) \mathbf{z}_{k-1, EMA}  \text{ with } \beta = \frac{2}{T-1}
\end{equation}
over all past observations of the object, where $T$ is the number of observed time steps of the respective instance. The observation $\mathbf{y}_{k}$ is used to update the Kalman filter. The optimal Kalman gain
\begin{equation}
   \mathbf{K}_{k} = \mathbf{P}_{k|k-1}\mathbf{H}^{T} ( \mathbf{H}\mathbf{P}_{k|k-1}\mathbf{H}^{T} + R)^{-1}
\end{equation}
is updated to minimize the residual error of the predicted model and the observation. The observation $\mathbf{y}_{k}$ is used to estimate the object state as
\begin{equation}
   \hat{\mathbf{x}}_{k|k} = \hat{\mathbf{x}}_{k|k-1} + \mathbf{K}_{k}(\mathbf{y}_{k} - \mathbf{H}\hat{\mathbf{x}}_{k|k-1}) 
\end{equation}
and with
\vspace*{-4pt}
\begin{equation}
   \mathbf{P}_{k} = \mathbf{P}_{k|k-1} - \mathbf{K}_k \mathbf{H}  \mathbf{P}_{k|k-1}
\end{equation}
the \textit{a posteriori} of the covariance matrix is updated.

\begin{table*}[t!]
\centering
\vspace*{-8pt}
\resizebox{\textwidth}{!}{
\begin{tabular}{l|l|cccc|c}
    \hline \hline
    \textbf{Training} & \textbf{Method} & \textbf{AMOTA} $\uparrow$ &	\textbf{AMOTP} (m) $\downarrow$ &	\textbf{Recall} $\uparrow$ & \textbf{MOTA} $\uparrow$ & \textbf{Modality} \\ 
    \textbf{Data Unseen} &  & &	& & & \\ 
        \hline
$\times$ & PF-Track~\cite{pang2023PFtrack} & 0.622 & 0.916 & 0.719 & 0.558 & Camera \\
$\times$ & QTrack ~\cite{yang2022qtrack} &  0.692 &  0.753 &  0.760 &  0.596 & Camera \\
$\times$ & QD-3DT ~\cite{hu2021QD3DT} & 0.425 &1.258 &  0.563 &  0.358 & Camera \\
\hline \hline
\textbf{\checkmark} & QD-3DT ~\cite{hu2021QD3DT} (trained on WOD) & 0.000 & 1.893 &  0.226 &  0.000 & Camera \\
$\times$ (CP) & CenterTrack~\cite{zhou2020CenterTrack} & 0.202 &1.195 &0.313 &0.134 & Camera \\
\textbf{\checkmark} (CP) & AB3DMOT ~\cite{weng2020AB3DMOT} & \underline{0.387} & \textbf{1.158} & \underline{0.506} & \underline{0.284} & Camera \\
\textbf{\checkmark} (CP) & Inverse Neural Rendering (ours) & \textbf{0.413} &  \underline{1.189} & \textbf{0.536} & \textbf{0.321} & Camera \\ 

\hline \hline
\end{tabular}
}
\caption{\textbf{Quantitative Evaluation for Camera-only Multi-Object Tracking}. Quantitative results on ``cars'' in the test split of the nuScenes tracking dataset~\cite{caesar2020nuscenes}. Our IR-based tracker outperforms AB3DMOT~\cite{weng2020AB3DMOT} on all metrics and CenterTrack~\cite{zhou2020CenterTrack} on accuracy. All three methods use the same detection backbone for fair comparison, while only CenterTrack requires end-to-end training on the dataset.
Additional results show on-par performance of our method with QD-3DT~\cite{hu2021QD3DT} trained on nuScenes \cite{caesar2020nuscenes}. QD-3DT trained on the Waymo Open Dataset (WOD) does not generalize to nuScenes and does not achieve competitive results. Only very recent transformer-based methods, such as PF-Track~\cite{pang2023PFtrack} and the metric learning approach of Q-Track~\cite{yang2022qtrack} achieve a higher score. However, these methods require end-to-end training on each dataset. ``CP'' denotes here the vision-only version of CenterPoint~\cite{zhou2019CenterPointVision} was used for object detection. \textbf{Bold} denotes best and \underline{underlined} second best for methods that did not train on the dataset or use the same detection backbone.}
\label{tab:nuScenes_results}
\end{table*}




\subsection{Implementation Details}\label{ssec:optim}
\noindent
\textbf{Representation Model.} 
We employ the GET3D~\cite{gao2022get3d} architecture as object model $G$. Following StyleGAN~\cite{karras2019styleGAN,karras2020styleGAN2} embeddings $z_{T}$ and $z_{S}$ are mapped to intermediate style embeddings $\mathbf{w}_S$ and $\mathbf{w}_T$ in a learned $\mathbf{W}\text{-space}$, which we optimize over instead of $\mathbf{Z}\text{-space}$. Style embeddings condition a generator function that produces tri-planes representing object shapes as Signed Distance Fields (SDFs) and textures as texture fields. We deliberately \emph{train our generator on synthetic data only}, see experiments below. Differentiable marching tetrahedra previously introduced in DMTet~\cite{shen2021dmtet} extract a mesh representation and Images are rendered with a differentiable rasterizer~\cite{laine2020modular}.

\vspace{0.5\baselineskip}
\noindent
\textbf{Computational Cost.}
Each IR optimization step in our implementation takes $\sim$ 0.3 seconds per frame. The generation and gradient computation through the generator determines the computational cost of the method. However, we note that the rendering pipeline, contributing the majority of the computational cost of the generator, has not been performance-optimized and can be naively parallelized when implemented in lower-level GL+CUDA graphics primitives.


\section{Experiments}\label{sec:exp}
In the following, we assess the proposed method. Having trained our generative scene model solely on simulated data~\cite{shapenet2015}, we test the generalization capabilities on the nuScenes \cite{caesar2020nuscenes} and Waymo~\cite{sun2020scalability} dataset -- both are unseen by the method. We analyze generative outputs of the test-time optimization and compare them against existing 3D multi-object trackers ~\cite{zhou2020CenterTrack, weng2020AB3DMOT, hu2021QD3DT,wang2023StreamPETR, yang2022qtrack} on camera-only data.

\subsection{Single-Shot Object Retrieval and Matching}
Although trained only on general object-centric synthetic data, ShapeNet~\cite{shapenet2015}, our method is capable of fitting a sample from the generative prior to observed objects in real datasets that match the vehicle type, color, and overall appearance closely, effectively making our method dataset-agnostic. 
We analyze the generations during optimization in the following.

\begin{figure*}[t!]
\centering
\resizebox{1.\linewidth}{!}{%
\begin{tabular}{@{}c@{\hskip .1cm}c@{\hskip .1cm}c@{\hskip .1cm}c@{\hskip .1cm}c@{}}

{Input Frame} & {Initial Guess} & {Texture Fitting} & {Object Pose Fitting} & {Shape Fitting} \\
	    
\includegraphics[width=.5\columnwidth, trim={0cm 0cm 0cm 0cm},clip]{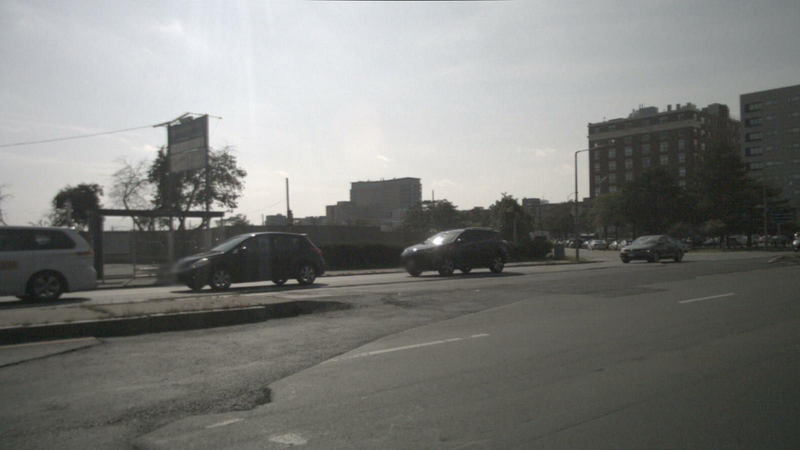}&
\includegraphics[width=.5\columnwidth, trim={0cm 0cm 0cm 0cm},clip]{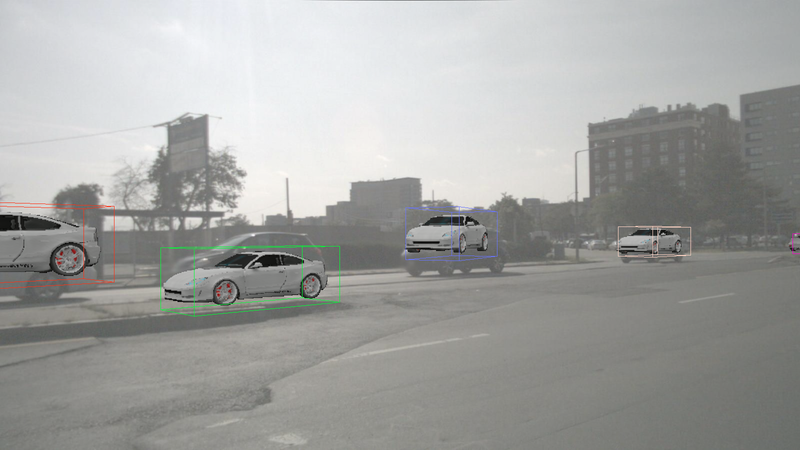}&
\includegraphics[width=.5\columnwidth, trim={0cm 0cm 0cm 0cm},clip]{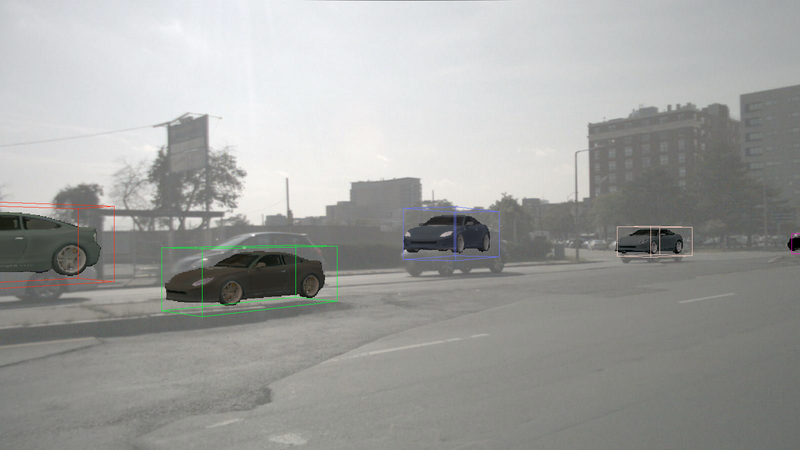}&
\includegraphics[width=.5\columnwidth, trim={0cm 0cm 0cm 0cm},clip]{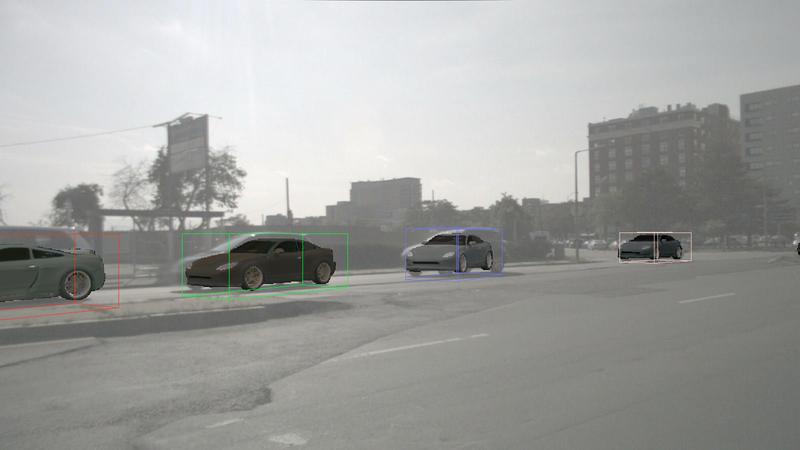}&
\includegraphics[width=.5\columnwidth, trim={0cm 0cm 0cm 0cm},clip]{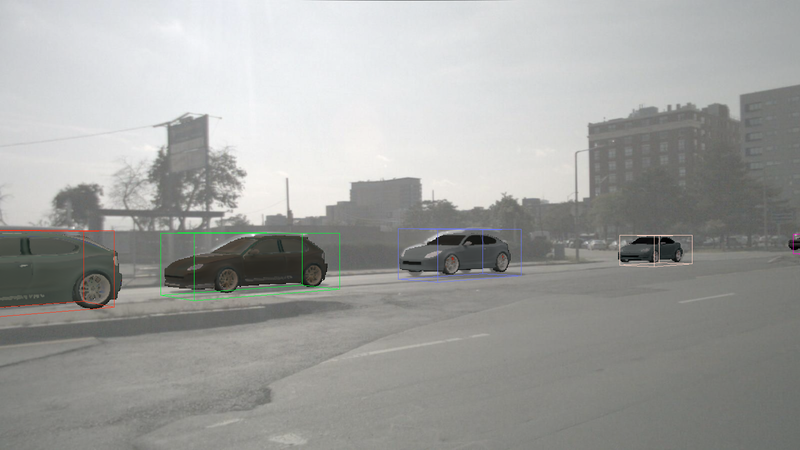}\\
& \multicolumn{4}{r}{\textit{Input frame is faded for visibility.}} \\
\end{tabular}}
\vspace*{-6pt} 
\caption{\textbf{Optimization Process.} From left to right, we show (i) the observed image, (ii) the rendering predicted by the initial starting point latent embeddings, 
(iii) the predicted rendered objects after the texture code is optimized (iv) the predicted rendered objects after the translation, scale, and rotation are optimized, and (v) the predicted rendered objects after the shape latent code is optimized. The ground truth images are faded to show our rendered objects clearly. Our method is capable of refining the predicted texture, pose, and shape over several optimization steps, even if initialized with poses or appearance far from the target -- all corrected through inverse rendering.} 
\label{fig:optim}
\end{figure*}

\vspace{0.5\baselineskip}
\noindent
\textbf{Optimization.} Given an image observation and coarse detections, our method aims to find the best 3D representation, including pose and appearance, solely through inverse rendering. In Fig.~\ref{fig:optim} we analyze this iterative optimization process, following a scheduled optimization as described in Sec.~\ref{ssec:optim}. We observe that the object's color is inferred in only two steps. Further, we can observe that even though the initial pose is incorrect, rotation and translation are optimized jointly through inverse rendering together with the shape and scale of the objects, recovering from sub-optimal initial guesses. The shape representation close to the observed object is reconstructed in just 5 steps. 



\subsection{Evaluation}\label{ssec:eval}
To provide a fair comparison of 3D multi-object tracking methods using monocular inputs, we compare against existing methods by running all our evaluations with the method reference code. We only evaluate methods, that consider past frames, but have no knowledge about future frames, which is a different task. While our method does not store the full history length of all images, we allow such memory techniques for other methods. We only consider purely mono-camera-based tracking methods.
We note that, in contrast to our method, most \emph{baseline methods we compare to are finetuned on the respective training set}. For all two-stage detect-and-track methods, we use CenterPoint~\cite{yin2021center} as the detection method. We compare to CenterTrack~\cite{zhou2020CenterTrack} as an established learning-based baseline, and present results of the very recent PFTrack~\cite{pang2023PFtrack}, a transformer-based tracking method, Qtrack~\cite{yang2022qtrack} as a metric learning method, and QD-3DT~\cite{hu2021QD3DT} as an LSTM-based state tracker combined with image feature matching. 
Of all learning-based methods, only CenterTrack~\cite{zhou2020CenterTrack} allows us to evaluate tracking performance with identical detections. Finally, we compare to AB3DMOT ~\cite{weng2020AB3DMOT} that builds on an arbitrary 3D detection algorithm and combines it with a modified Kalman filter to track the state of each object. AB3DMOT~\cite{weng2020AB3DMOT} and the proposed method are the only methods that are data-agnostic in the sense that they have not seen the training dataset. For a fair evaluation of these generalization capabilities in learning-based methods, we include another version of QD-3DT solely trained on the Waymo Open Dataset~\cite{sun2020scalability} and evaluate on nuScenes~\cite{caesar2020nuscenes}.  We discuss the findings in the following.


\vspace{0.5\baselineskip}
\noindent
\textbf{Validation on nuScenes.} Tab.~\ref{tab:nuScenes_results} reports quantitative results on the test split of the nuScenes tracking dataset~\cite{caesar2020nuscenes} on the car object class for all six cameras. We list results for the multi-object tracking accuracy (MOTA)~\cite{bernardin2006MOTA} metric, the AMOTA~\cite{weng2020AB3DMOT} metric, average multi-object tracking precision (AMOTP)~\cite{weng2020AB3DMOT} and recall of all methods. First, we evaluate a version of QD-3DT~\cite{hu2021QD3DT} that has been trained on the Waymo Open Dataset~\cite{sun2020scalability} (WOD) but tested on nuScenes. This experiment is reported in row four of Tab.~\ref{tab:nuScenes_results} and confirms that recent end-to-end detection and tracking methods do not perform well on unseen data (see qualitative results in the Supplementary Material). Moreover, perhaps surprisingly, even when using use the same vision-only detection backbone as in our approach, the established end-to-end trained baseline CenterTrack~\cite{zhou2020CenterTrack}, which has seen the dataset, performs worse than our method.
Our IR-based method outperforms the general tracker AB3DMOT~\cite{weng2020AB3DMOT}. 
When other methods are given access to the dataset, recent learning-based methods such as the end-to-end LSTM-based method QD-3DT~\cite{hu2021QD3DT} perform on par.  Only the most recent transformer-based methods such as PF-Track~\cite{pang2023PFtrack} and the QTrack~\cite{yang2022qtrack}, which employ a quality-based association model on a large set of learned metrics, such as heat maps and depth, achieve higher scores. Note again, that these methods, in contrast to the proposed method, have been trained on this dataset and cannot be evaluated independently of their detector performance.

\begin{figure}[t!]
	 \renewcommand{\arraystretch}{0.4}
	\centering
	\resizebox{1\columnwidth}{!}{%
	\begin{tabular}{@{}c@{\hskip .1cm}c@{\hskip .1cm}c@{}}
 
    \multicolumn{1}{c}{\large (i) Input Frame} & \multicolumn{1}{c}{\large (ii) INR 3D Generation} & \multicolumn{1}{c}{\large (iii) INR BEV Layout} \\
     
    \includegraphics[width=.5\columnwidth, trim={0cm 0cm 0cm 0cm},clip]{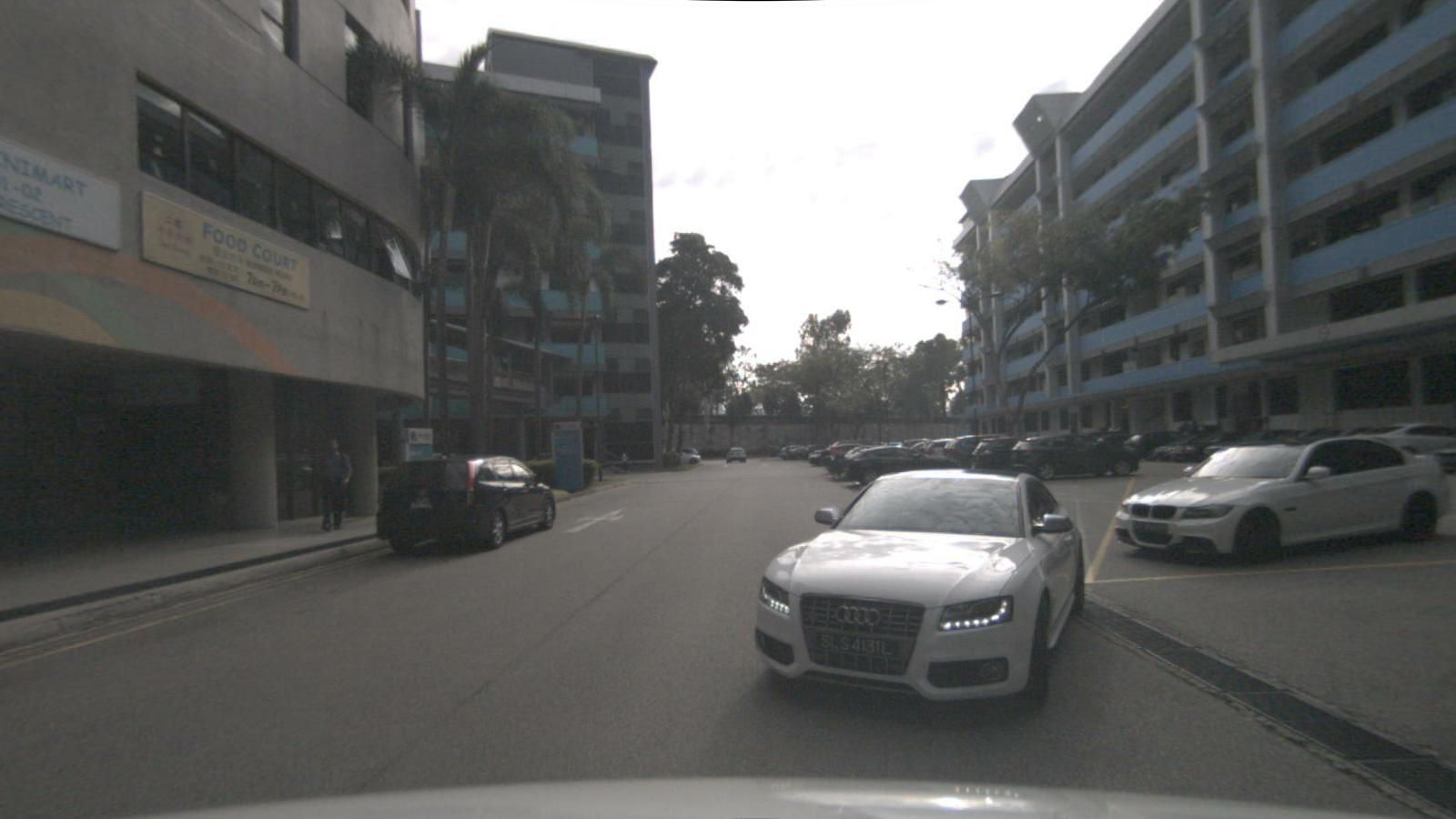}&

    \includegraphics[width=.5\columnwidth, trim={0cm 0cm 0cm 0cm},clip]{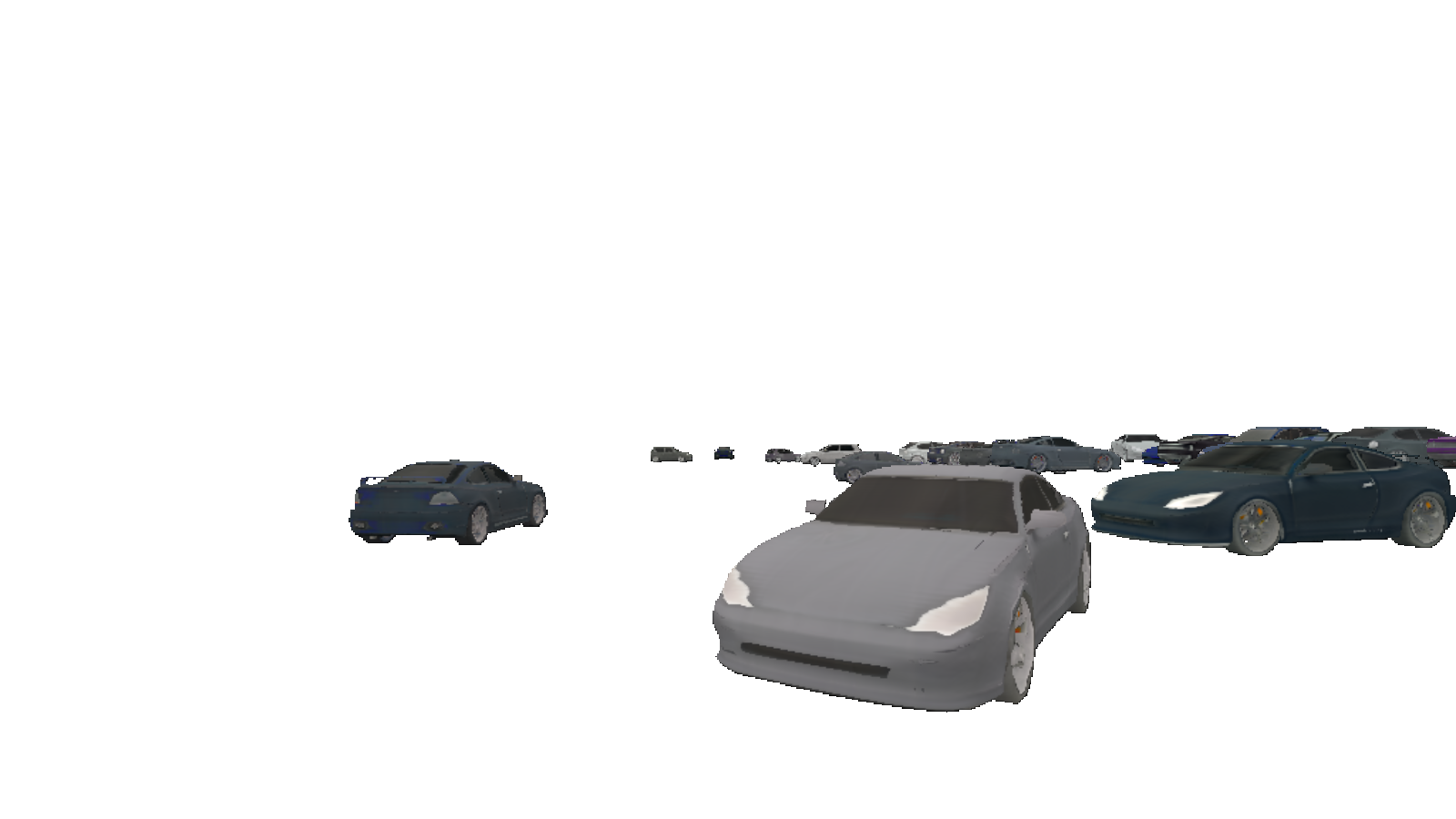}&

    \includegraphics[width=.5\columnwidth, trim={2.1cm 2cm 1.9cm 2cm},clip]{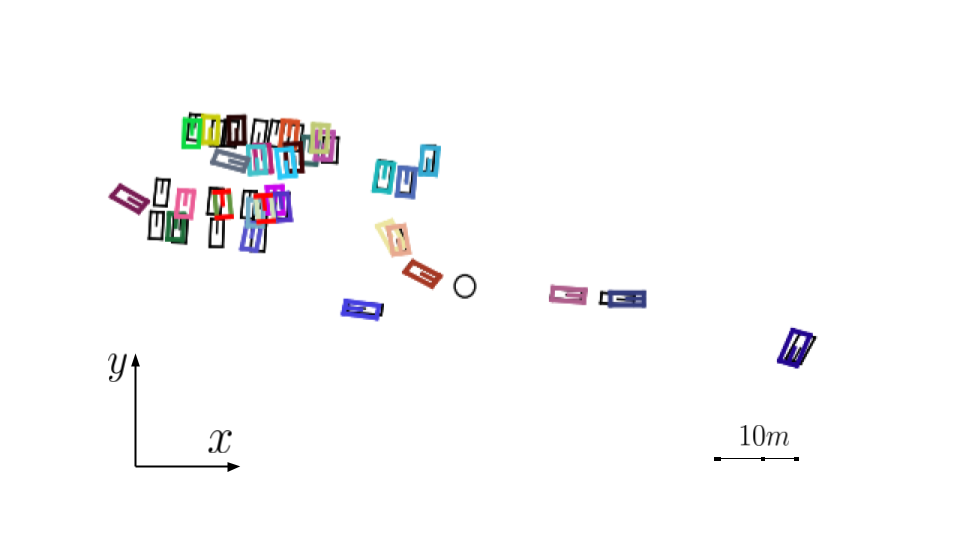}\\

    \includegraphics[width=.5\columnwidth, trim={26cm 6.7cm 12.5cm 15cm},clip]{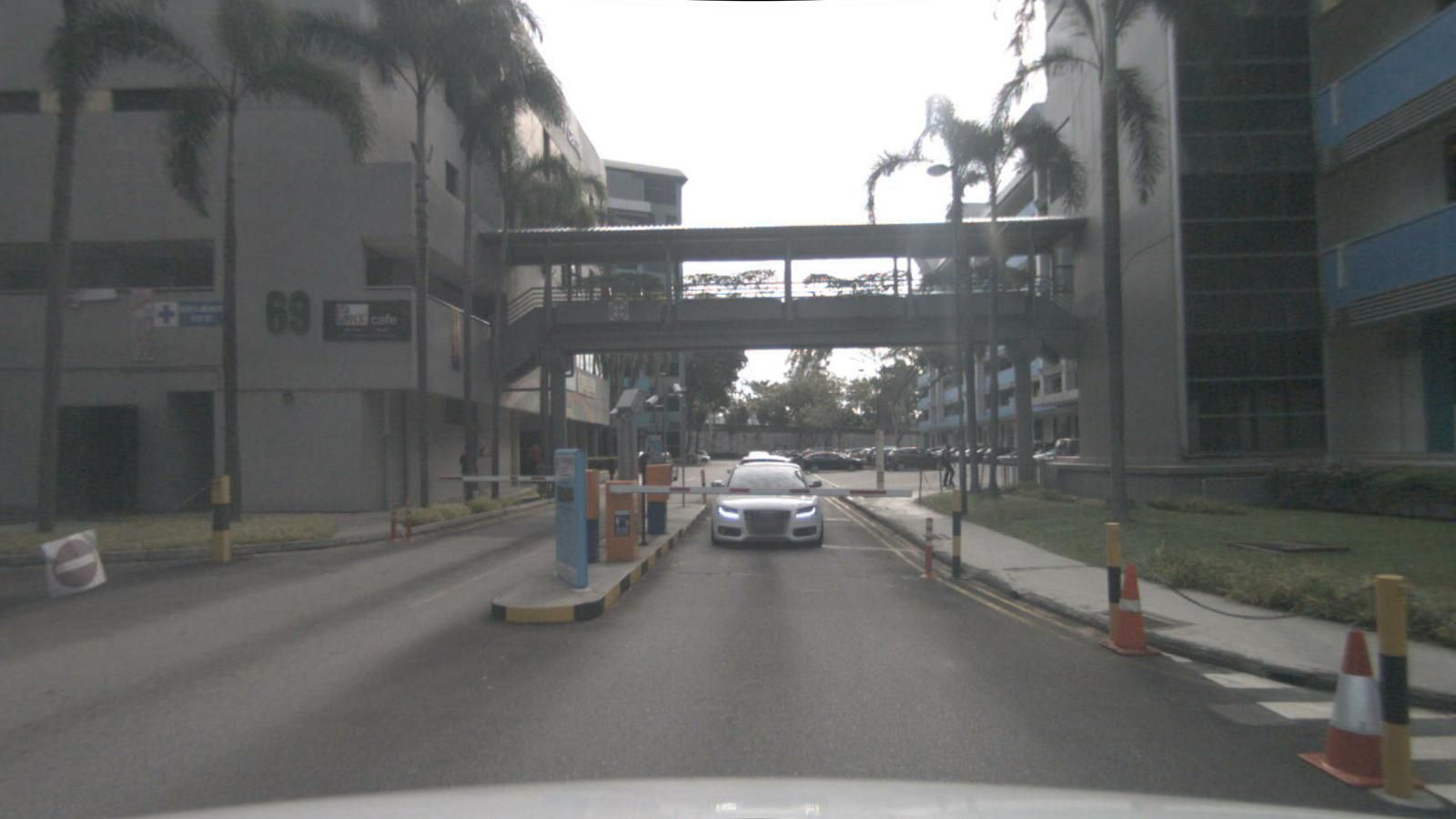}&
    \includegraphics[width=.5\columnwidth, trim={26cm 6.7cm 12.5cm 15cm},clip]{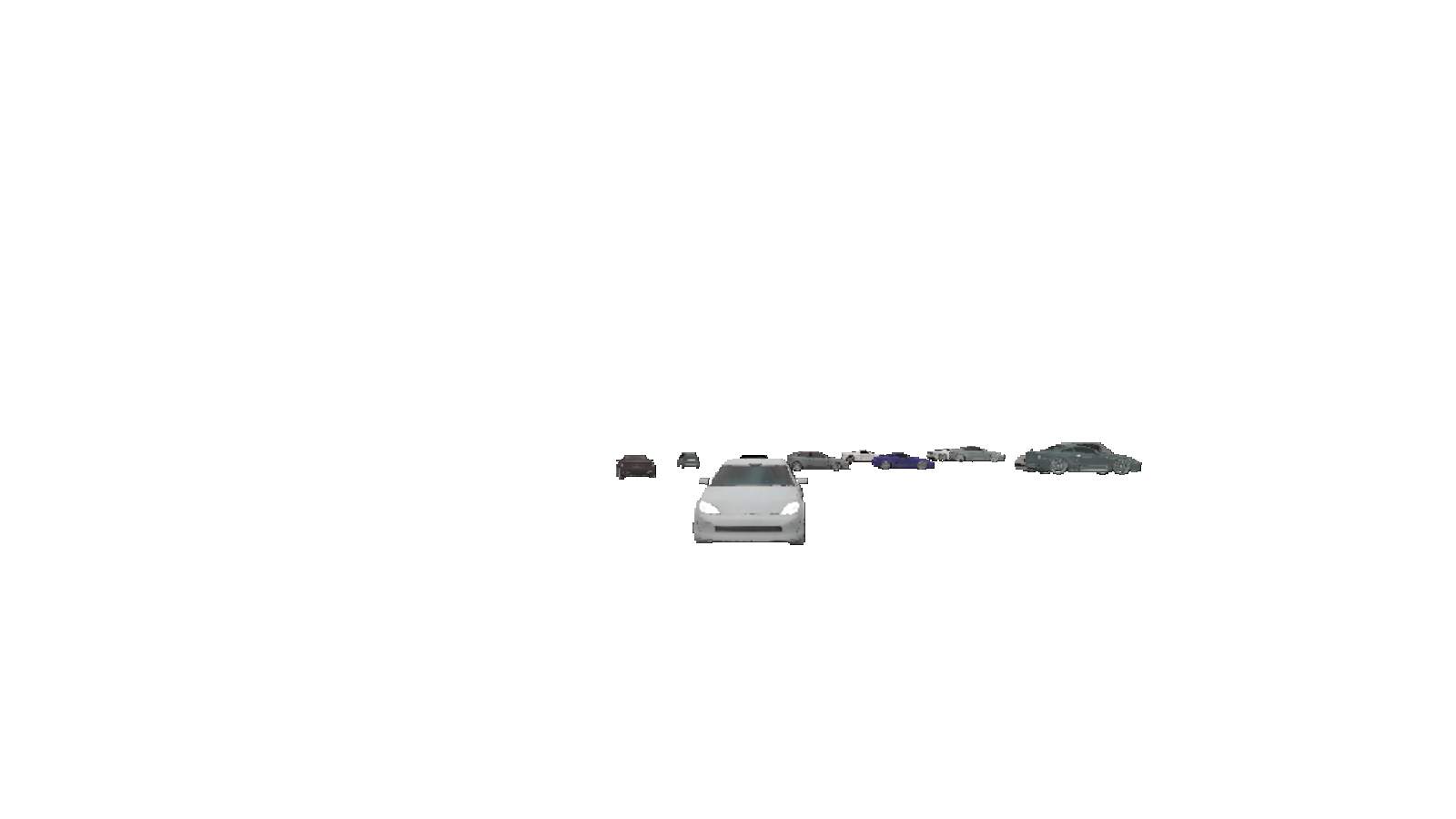}&
    \includegraphics[width=.5\columnwidth, trim={2.1cm 2cm 1.9cm 2cm},clip]{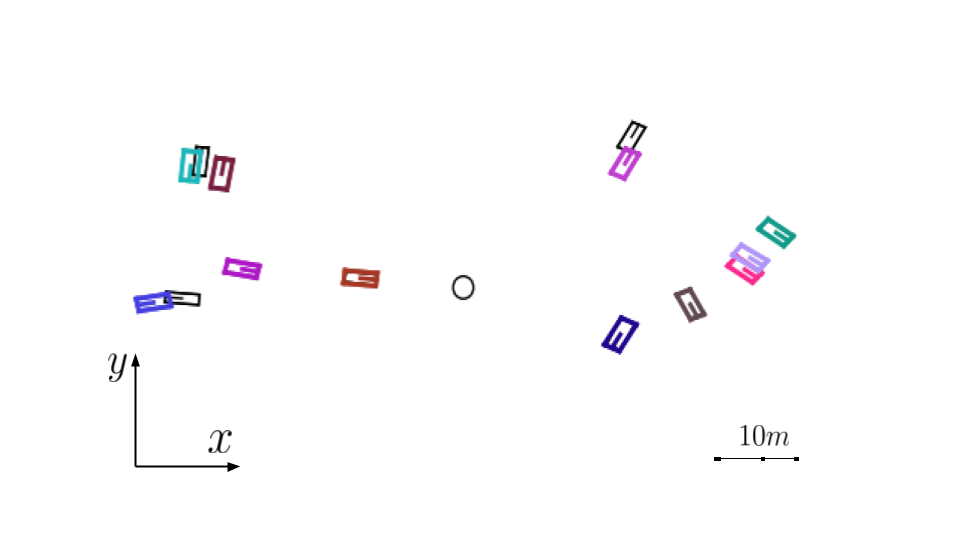}\\
     
    \end{tabular}}
\caption{\textbf{Layout Generation Through Inverse Rendering.} From left to right, we show (i) observed image from a single camera for two scenes, (ii) test-time optimized inverse rendered (IR) objects of class ``car'', and (iii) Bird's Eye View (BEV) layout of the scene. In the BEV layout, black boxes represent ground truth, and the colored boxes represent predicted BEV boxes. The bottom shows a zoomed-in region at a 60 m distance (see BEV layout). Even in this setting, our method recovers the coarse appearance, shape of the objects, pose, and size,} 
\label{fig:BEV3D_fig}
\end{figure}

We visualize the rendered objects predicted by our tracking method in Fig.~\ref{fig:nuScenes_results}. We show an observed image from a single camera at time step $k = 0$, followed by rendered objects overlaid over the observed image at time step $k = 0, 1, 2 \text{ and } 3$ along with their respective bounding boxes, with color-coded tracklets. We see that our method does not lose any tracks in challenging scenarios in diverse scenes shown here, from dense urban areas to suburban traffic crossings, and handles occlusions and clutter effectively. By visualizing the rendered objects as well as analyzing the loss values, our method allows us to reason about and explain success and failure cases effectively, enabling explainable 3D object tracking. The rendered output images provide interpretable inference results that explain successful or failed matching due to shadows, appearance, shape, or pose. For example, the blue car in the IR inference in Fig.~\ref{fig:BEV3D_fig} top row was incorrectly matched due to an appearance mismatch in a shadow region. A rendering model including ambient illumination may resolve this ambiguity, see further discussion in the Supplementary Material.

\vspace{0.5\baselineskip}
\noindent
\textbf{Interpretation.} Fig~\ref{fig:BEV3D_fig} shows the inverse rendered scene graphs in isolation and birds-eye-view tracking outputs on a layout level. Our method accurately recovers the object pose, instance type, appearance, and scale. As such, our approach directly outputs a 3D model of the full scene, i.e., layout and object instances, along with the temporal history of the scene recovered through tracking -- a rich scene representation that can be directly ingested by downstream planning and control tasks, or simulation methods to train downstream tasks. As such, the method also allows us to reason about the scene by leveraging the 3D information provided by our predicted 3D representations. The 3D locations, object orientations, and sizes recovered from such visualizations can not only enable us to explain the predictions of our object tracking method, especially in the presence of occlusions or ID switches but also be used in other downstream tasks that require rich 3D understanding, such as planning. 

\vspace{0.5\baselineskip}
\noindent
\textbf{Validation on Waymo.}
Next, we provide qualitative results from the 3D tracking on the validation set of WOD~\cite{sun2020scalability} in Fig.~\ref{fig:waymo_results}. The \emph{only public results} on the provided test set are presented in QD-3DT~\cite{hu2021QD3DT}, which may indicate it fails on this dataset. While the size of the dataset and its variety is of high interest for all autonomous driving tasks, Hu et al.~\cite{hu2021QD3DT} conclude that vision-only test set evaluation is not representative of a test set developed for surround view lidar data on partial unobserved camera images only. As such, we provide here qualitative results in Fig.~\ref{fig:waymo_results}, which validate that the method achieves tracking of similar quality on all datasets, providing a generalizing tracking approach. We show that our method does not lose tracks on Waymo scenes in diverse conditions.

\begin{table}[t]
\caption{\textbf{Ablation Experiments.}}

\begin{subtable}[t]{1.\linewidth}
\centering
\renewcommand{\arraystretch}{1.0}
\centering
\resizebox{1.\textwidth}{!}{
\begin{tabular}{@{}c@{\hskip .1cm}c@{\hskip .1cm}c@{}}

    \multicolumn{1}{c}{(a) Input Frame} &
    \multicolumn{1}{c}{(b) Full (ours)} & 
    \multicolumn{1}{c}{(c) \underline{No Schedule}} \\
    
    \includegraphics[width=.3\columnwidth, trim={0cm 0cm 0cm 0cm},clip]{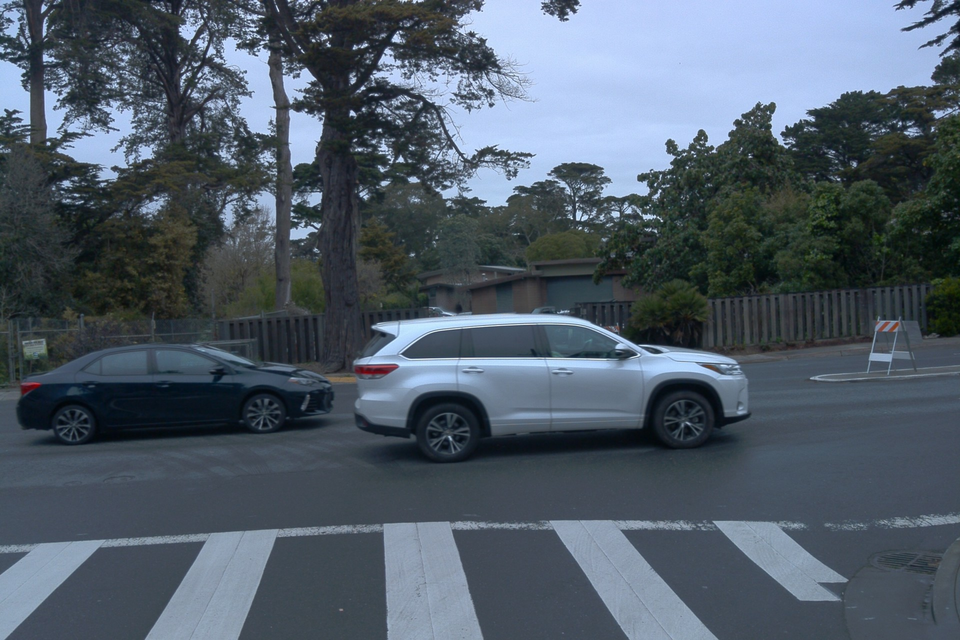}&
    \includegraphics[width=.3\columnwidth, trim={0cm 0cm 0cm 0cm},clip]{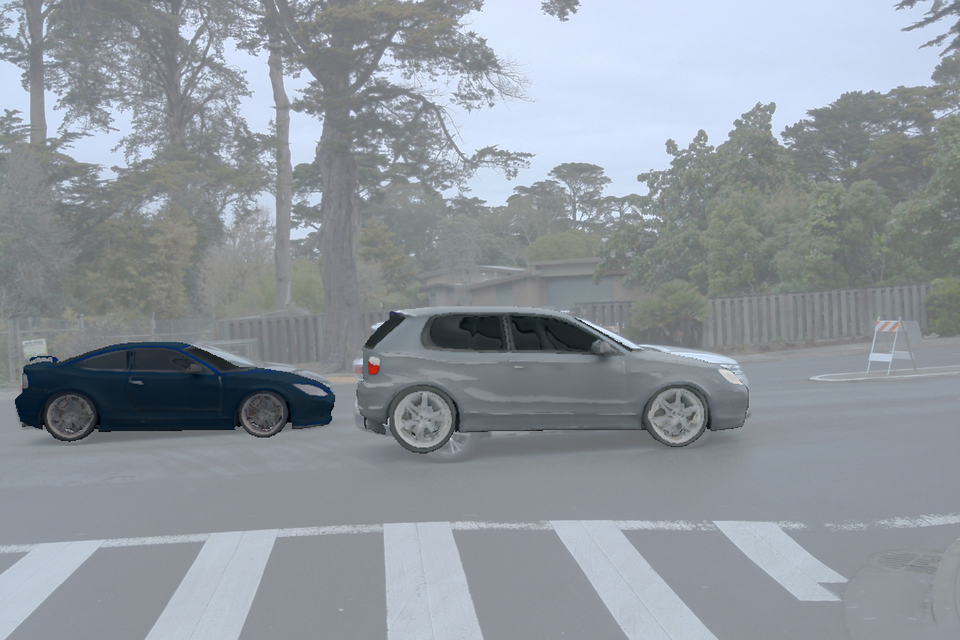}&
    \includegraphics[width=.3\columnwidth, trim={0cm 0cm 0cm 0cm},clip]{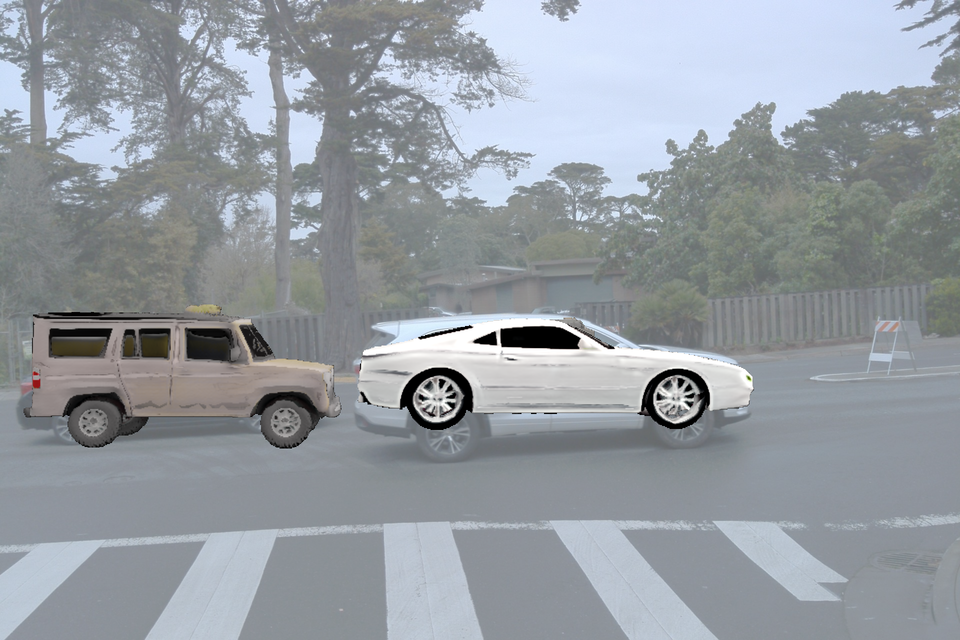} \\

    \includegraphics[width=.3\columnwidth, trim={11cm 8.34cm 15cm 9cm},clip]{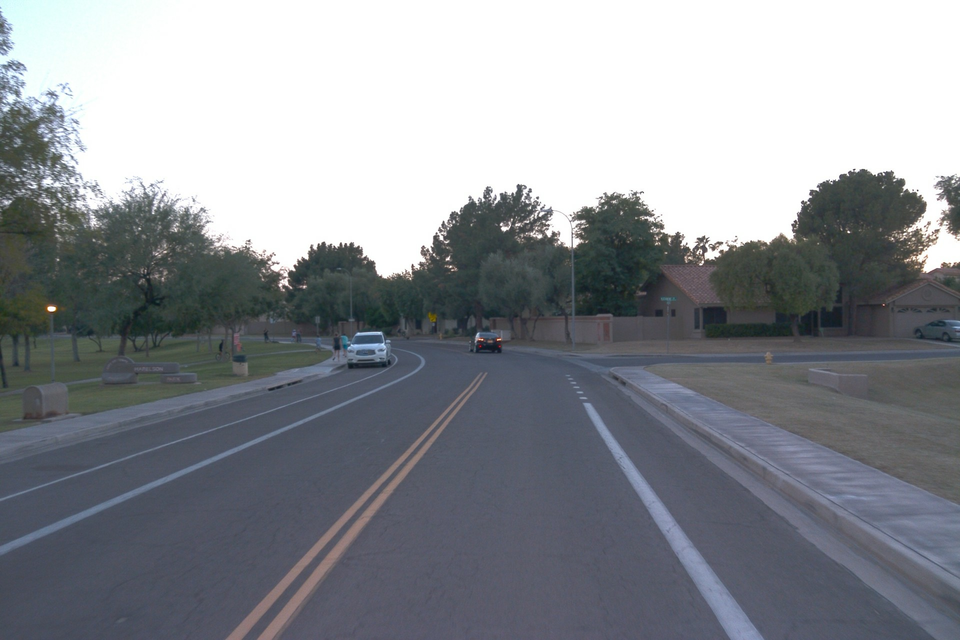} &
    \includegraphics[width=.3\columnwidth, trim={11cm 8.34cm 15cm 9cm},clip]{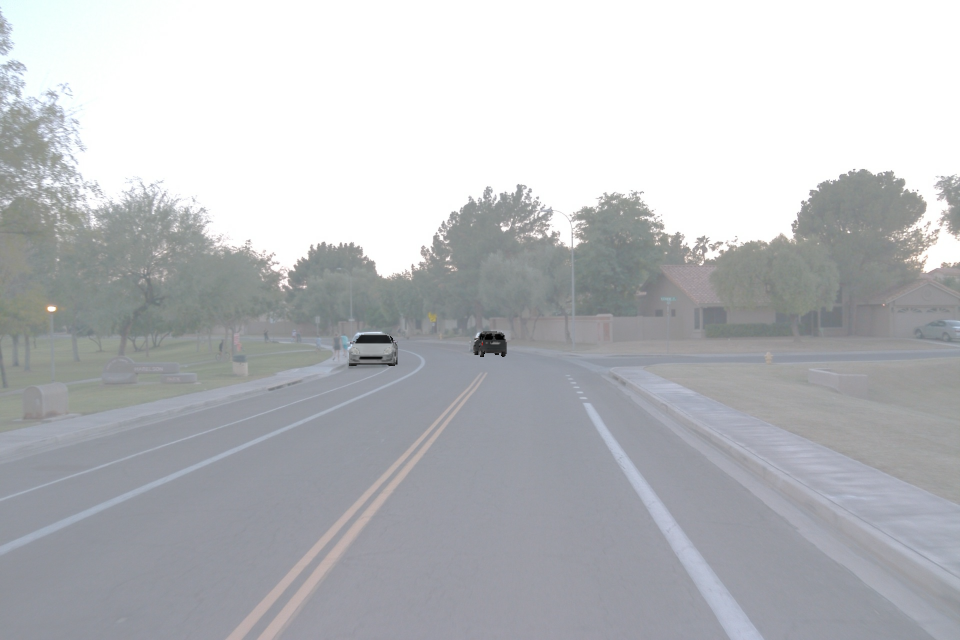} &
    \includegraphics[width=.3\columnwidth, trim={11cm 8.34cm 15cm 9cm},clip]{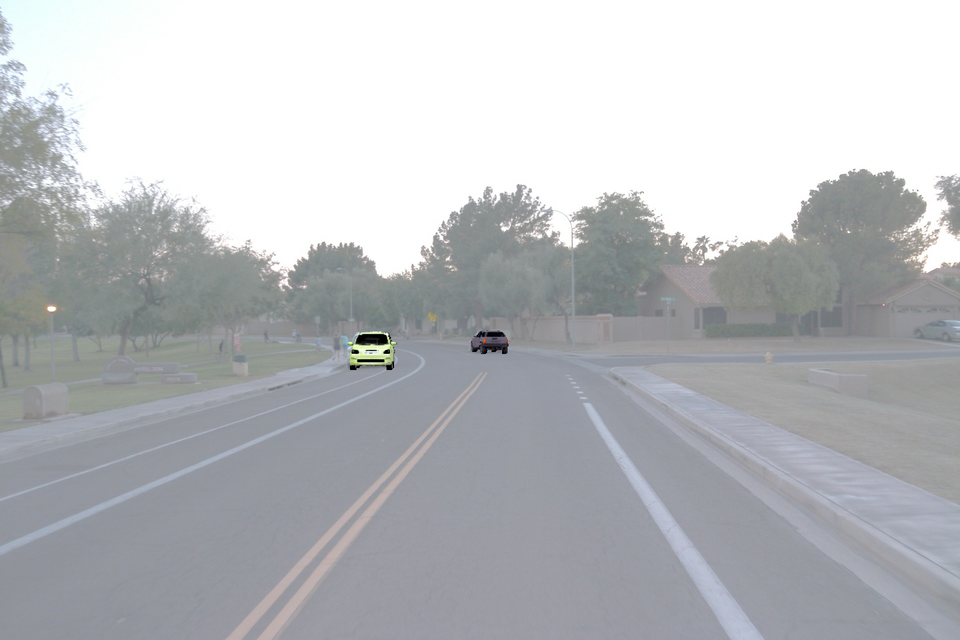} \\
    
    \multicolumn{3}{r}{\textit{Input frame is faded for visibility.}}
    \end{tabular}
}

\caption{\textbf{Effect of Optimization Schedule.} (a) observed image, (b) optimized generations using the proposed schedule in Sec.~\ref{sec:method}, (c) optimized generations using no schedule. This supports the quantitative to the left.}
\label{fig:opt_scheduler}
\end{subtable}
\vspace*{6pt}

\begin{subtable}[t]{1.\linewidth}
\centering
\resizebox{1.\textwidth}{!}{
\begin{tabular}{l|lll}
    \hline
    \hline
    Method & AMOTA $\uparrow$ & Recall $\uparrow$  & MOTA$\uparrow$  \\
    \hline
    $\mathcal{L}_{IR}$ \& $\mathcal{L}_{embed}$ - Eq.\ref{eq:regularize}  & \textbf{0.112} & \textbf{0.264 } & \textbf{0.113} \\
    $\mathcal{L}_{IR}$ - Eq.\ref{eq:loss_mse_lpips}  & 0.103 & 0.236   & 0.112 \\
    $\mathcal{L}_{perceptual}$ - Eq.~\ref{eq:loss_lpips} & 0.100 & 0.251 & 0.101 \\
    $\mathcal{L}_{RGB}$ - Eq.~\ref{eq:loss_mse} & N/A  & N/A & N/A  \\
    \hline
    \underline{No Schedule} & 0.102 & 0.224 & 0.110  \\
\hline
\hline
\end{tabular}
}
\caption{\textbf{Optimization Schedule and Loss Components.} Ablations were run on a small subset of the nuScenes~\cite{caesar2020nuscenes} validation set.  $\mathcal{L}_{RGB}$ fails due to the optimizer fitting objects to the background instead, increasing the size of each object resulting in out of memory.}
\label{tab:optim_ablations}
\end{subtable}
\end{table}

\vspace{0.5\baselineskip}
\noindent
\subsection{Ablation Experiments}
As ablation experiments, we analyze the optimization schedule, the INR loss function components, and the weights of the tracker, applying them to a subset of scenes from the nuScenes validation set. We deliberately select this smaller validation set due to its increased difficulty.
The top row of Tab.~\ref{tab:optim_ablations} lists the quantitative results from our ablation study of the optimization scheduler. Our findings reveal a crucial insight: the strength of our method lies not in isolated loss components but in their synergistic integration. Specifically, the amalgamation of pixel-wise, perceptual, and embedding terms significantly enhances AMOTA, MOTA, and Recall metrics.

Moreover, the absence of an optimization schedule led to less robust matching as quantitative and qualitative results in Tab.~\ref{fig:opt_scheduler} reveal.  However, the core efficacy of our tracking method remained intact as indicated in the last row of Tab.~\ref{tab:optim_ablations}. This nuanced understanding underscores the importance of component interplay in our method.




%
\section{Conclusion}
We investigate inverse neural rendering as an alternative to existing feed-forward tracking methods. Specifically, we recast 3D multi-object tracking from RGB cameras as an inverse test-time optimization problem over the latent space of pre-trained 3D object representations that, when rendered, best represent object instances in a given input image. We optimize an image loss over generative latent spaces that inherently disentangle shape and appearance properties. This approach to tracking also enables examining the reconstructed objects, reasoning about failure situations, and resolving ambiguous cases -- rendering object layouts and loss function values provides interpretability ``for free''. We validate that the method has high generalization capabilities, and without seeing a dataset, outperforms existing tracking methods' generalization capabilities. In the future, we hope to investigate not only object detection with inverse rendering but broad, in-the-wild object class identification via conditional generation methods -- unlocking analysis-by-synthesis in vision with generative neural rendering.


%
%
\bibliographystyle{ieeetr}
\bibliography{bib}

\begin{thebibliography}{10}

\bibitem{long2015fully}
J.~Long, E.~Shelhamer, and T.~Darrell, ``Fully convolutional networks for
  semantic segmentation,'' in {\em Proceedings of the IEEE conference on
  computer vision and pattern recognition}, pp.~3431--3440, 2015.

\bibitem{li2017fully}
Y.~Li, H.~Qi, J.~Dai, X.~Ji, and Y.~Wei, ``Fully convolutional instance-aware
  semantic segmentation,'' in {\em Proceedings of the IEEE conference on
  computer vision and pattern recognition}, pp.~2359--2367, 2017.

\bibitem{chen2014semantic}
L.-C. Chen, G.~Papandreou, I.~Kokkinos, K.~Murphy, and A.~L. Yuille, ``Semantic
  image segmentation with deep convolutional nets and fully connected crfs,''
  {\em arXiv preprint arXiv:1412.7062}, 2014.

\bibitem{ren2015faster}
S.~Ren, K.~He, R.~Girshick, and J.~Sun, ``Faster r-cnn: Towards real-time
  object detection with region proposal networks,'' {\em Advances in neural
  information processing systems}, vol.~28, 2015.

\bibitem{girshick2015fast}
R.~Girshick, ``Fast r-cnn,'' in {\em Proceedings of the IEEE international
  conference on computer vision}, pp.~1440--1448, 2015.

\bibitem{redmon2016yolo}
J.~Redmon, S.~Divvala, R.~Girshick, and A.~Farhadi, ``You only look once:
  Unified, real-time object detection,'' in {\em Proceedings of the IEEE
  conference on computer vision and pattern recognition}, pp.~779--788, 2016.

\bibitem{liu2016ssd}
W.~Liu, D.~Anguelov, D.~Erhan, C.~Szegedy, S.~Reed, C.-Y. Fu, and A.~C. Berg,
  ``Ssd: Single shot multibox detector,'' in {\em European conference on
  computer vision}, pp.~21--37, Springer, 2016.

\bibitem{ku2018joint}
J.~Ku, M.~Mozifian, J.~Lee, A.~Harakeh, and S.~L. Waslander, ``Joint 3d
  proposal generation and object detection from view aggregation,'' in {\em
  2018 IEEE/RSJ International Conference on Intelligent Robots and Systems
  (IROS)}, pp.~1--8, IEEE, 2018.

\bibitem{qi2018frustum}
C.~R. Qi, W.~Liu, C.~Wu, H.~Su, and L.~J. Guibas, ``Frustum pointnets for 3d
  object detection from rgb-d data,'' in {\em Proceedings of the IEEE
  conference on computer vision and pattern recognition}, pp.~918--927, 2018.

\bibitem{zhou2018voxelnet}
Y.~Zhou and O.~Tuzel, ``Voxelnet: End-to-end learning for point cloud based 3d
  object detection,'' in {\em Proceedings of the IEEE conference on computer
  vision and pattern recognition}, pp.~4490--4499, 2018.

\bibitem{sharma2018beyond}
S.~Sharma, J.~A. Ansari, J.~K. Murthy, and K.~M. Krishna, ``Beyond pixels:
  Leveraging geometry and shape cues for online multi-object tracking,'' in
  {\em 2018 IEEE International Conference on Robotics and Automation (ICRA)},
  pp.~3508--3515, IEEE, 2018.

\bibitem{kim2021eagermot}
A.~Kim, A.~O{\v{s}}ep, and L.~Leal-Taix{\'e}, ``Eagermot: 3d multi-object
  tracking via sensor fusion,'' in {\em 2021 IEEE International Conference on
  Robotics and Automation (ICRA)}, pp.~11315--11321, IEEE, 2021.

\bibitem{zhou2020CenterTrack}
X.~Zhou, V.~Koltun, and P.~Kr{\"a}henb{\"u}hl, ``Tracking objects as points,''
  in {\em European Conference on Computer Vision}, pp.~474--490, Springer,
  2020.

\bibitem{chaabane2021deft}
M.~Chaabane, P.~Zhang, J.~R. Beveridge, and S.~O'Hara, ``Deft: Detection
  embeddings for tracking,'' {\em arXiv preprint arXiv:2102.02267}, 2021.

\bibitem{yin2021center}
T.~Yin, X.~Zhou, and P.~Krahenbuhl, ``Center-based 3d object detection and
  tracking,'' in {\em Proceedings of the IEEE/CVF conference on computer vision
  and pattern recognition}, pp.~11784--11793, 2021.

\bibitem{weng2020AB3DMOT}
X.~Weng, J.~Wang, D.~Held, and K.~Kitani, ``3d multi-object tracking: A
  baseline and new evaluation metrics,'' in {\em 2020 IEEE/RSJ International
  Conference on Intelligent Robots and Systems (IROS)}, pp.~10359--10366, IEEE,
  2020.

\bibitem{pang2021simpletrack}
Z.~Pang, Z.~Li, and N.~Wang, ``Simpletrack: Understanding and rethinking 3d
  multi-object tracking,'' {\em arXiv preprint arXiv:2111.09621}, 2021.

\bibitem{wang2019densefusion}
C.~Wang, D.~Xu, Y.~Zhu, R.~Mart{\'\i}n-Mart{\'\i}n, C.~Lu, L.~Fei-Fei, and
  S.~Savarese, ``Densefusion: 6d object pose estimation by iterative dense
  fusion,'' in {\em Proceedings of the IEEE/CVF conference on computer vision
  and pattern recognition}, pp.~3343--3352, 2019.

\bibitem{xiang2017posecnn}
Y.~Xiang, T.~Schmidt, V.~Narayanan, and D.~Fox, ``Posecnn: A convolutional
  neural network for 6d object pose estimation in cluttered scenes,'' {\em
  arXiv preprint arXiv:1711.00199}, 2017.

\bibitem{liu2022bevfusion}
Z.~Liu, H.~Tang, A.~Amini, X.~Yang, H.~Mao, D.~Rus, and S.~Han, ``Bevfusion:
  Multi-task multi-sensor fusion with unified bird's-eye view representation,''
  {\em arXiv}, 2022.

\bibitem{weng2020gnn3dmot}
X.~Weng, Y.~Wang, Y.~Man, and K.~M. Kitani, ``Gnn3dmot: Graph neural network
  for 3d multi-object tracking with 2d-3d multi-feature learning,'' in {\em
  Proceedings of the IEEE/CVF Conference on Computer Vision and Pattern
  Recognition}, pp.~6499--6508, 2020.

\bibitem{focalformer3d}
Y.~Chen, Z.~Yu, Y.~Chen, S.~Lan, A.~Anandkumar, J.~Jia, and J.~M. Alvarez,
  ``Focalformer3d: Focusing on hard instance for 3d object detection,'' in {\em
  Proceedings of the IEEE/CVF International Conference on Computer Vision},
  pp.~8394--8405, 2023.

\bibitem{bai2021pointdsc}
B.~Xuyang, Z.~Hu, X.~Zhu, Q.~Huang, Y.~Chen, H.~Fu, and C.-L. Tai,
  ``{TransFusion}: {R}obust {L}idar-{C}amera {F}usion for {3}d {O}bject
  {D}etection with {T}ransformers,'' {\em CVPR}, 2022.

\bibitem{hu2021QD3DT}
H.-N. Hu, Y.-H. Yang, T.~Fischer, T.~Darrell, F.~Yu, and M.~Sun, ``Monocular
  quasi-dense 3d object tracking,'' {\em IEEE Transactions on Pattern Analysis
  and Machine Intelligence}, 2022.

\bibitem{wu2021trades}
J.~Wu, J.~Cao, L.~Song, Y.~Wang, M.~Yang, and J.~Yuan, ``Track to detect and
  segment: An online multi-object tracker,'' in {\em Proceedings of the
  IEEE/CVF conference on computer vision and pattern recognition},
  pp.~12352--12361, 2021.

\bibitem{marinello2022triplettrack}
N.~Marinello, M.~Proesmans, and L.~Van~Gool, ``Triplettrack: 3d object tracking
  using triplet embeddings and lstm,'' in {\em Proceedings of the IEEE/CVF
  Conference on Computer Vision and Pattern Recognition}, pp.~4500--4510, 2022.

\bibitem{nguyen2022multiCamMultiTrack}
P.~Nguyen, K.~G. Quach, C.~N. Duong, N.~Le, X.-B. Nguyen, and K.~Luu,
  ``Multi-camera multiple 3d object tracking on the move for autonomous
  vehicles,'' in {\em Proceedings of the IEEE/CVF Conference on Computer Vision
  and Pattern Recognition}, pp.~2569--2578, 2022.

\bibitem{gladkova2022directtracker}
M.~Gladkova, N.~Korobov, N.~Demmel, A.~O{\v{s}}ep, L.~Leal-Taix{\'e}, and
  D.~Cremers, ``Directtracker: 3d multi-object tracking using direct image
  alignment and photometric bundle adjustment,'' {\em arXiv preprint
  arXiv:2209.14965}, 2022.

\bibitem{yang2022qtrack}
J.~Yang, E.~Yu, Z.~Li, X.~Li, and W.~Tao, ``Quality matters: Embracing quality
  clues for robust 3d multi-object tracking,'' {\em arXiv preprint
  arXiv:2208.10976}, 2022.

\bibitem{pang2023PFtrack}
Z.~Pang, J.~Li, P.~Tokmakov, D.~Chen, S.~Zagoruyko, and Y.-X. Wang, ``Standing
  between past and future: Spatio-temporal modeling for multi-camera 3d
  multi-object tracking,'' in {\em Proceedings of the IEEE/CVF Conference on
  Computer Vision and Pattern Recognition}, 2023.

\bibitem{wang2023StreamPETR}
S.~Wang, Y.~Liu, T.~Wang, Y.~Li, and X.~Zhang, ``Exploring object-centric
  temporal modeling for efficient multi-view 3d object detection,'' {\em arXiv
  preprint arXiv:2303.11926}, 2023.

\bibitem{gao2022get3d}
J.~Gao, T.~Shen, Z.~Wang, W.~Chen, K.~Yin, D.~Li, O.~Litany, Z.~Gojcic, and
  S.~Fidler, ``Get3d: A generative model of high quality 3d textured shapes
  learned from images,'' in {\em Advances In Neural Information Processing
  Systems}, 2022.

\bibitem{park2019deepsdf}
J.~J. Park, P.~Florence, J.~Straub, R.~Newcombe, and S.~Lovegrove, ``Deepsdf:
  Learning continuous signed distance functions for shape representation,'' in
  {\em Proceedings of the IEEE Conference on Computer Vision and Pattern
  Recognition (CVPR)}, June 2019.

\bibitem{mildenhall2020nerf}
B.~Mildenhall, P.~P. Srinivasan, M.~Tancik, J.~T. Barron, R.~Ramamoorthi, and
  R.~Ng, ``Nerf: Representing scenes as neural radiance fields for view
  synthesis,'' 2020.

\bibitem{shen2023gina3d}
B.~Shen, X.~Yan, C.~R. Qi, M.~Najibi, B.~Deng, L.~Guibas, Y.~Zhou, and
  D.~Anguelov, ``Gina-3d: Learning to generate implicit neural assets in the
  wild,'' in {\em Proceedings of the IEEE/CVF Conference on Computer Vision and
  Pattern Recognition (CVPR)}, pp.~4913--4926, June 2023.

\bibitem{caesar2020nuscenes}
H.~Caesar, V.~Bankiti, A.~H. Lang, S.~Vora, V.~E. Liong, Q.~Xu, A.~Krishnan,
  Y.~Pan, G.~Baldan, and O.~Beijbom, ``nuscenes: A multimodal dataset for
  autonomous driving,'' in {\em Proceedings of the IEEE/CVF conference on
  computer vision and pattern recognition}, pp.~11621--11631, 2020.

\bibitem{sun2020scalability}
P.~Sun, H.~Kretzschmar, X.~Dotiwalla, A.~Chouard, V.~Patnaik, P.~Tsui, J.~Guo,
  Y.~Zhou, Y.~Chai, B.~Caine, {\em et~al.}, ``Scalability in perception for
  autonomous driving: Waymo open dataset,'' in {\em Proceedings of the IEEE/CVF
  Conference on Computer Vision and Pattern Recognition (CVPR)},
  pp.~2446--2454, 2020.

\bibitem{yilmaz2006object}
A.~Yilmaz, O.~Javed, and M.~Shah, ``Object tracking: A survey,'' {\em Acm
  computing surveys (CSUR)}, vol.~38, no.~4, pp.~13--es, 2006.

\bibitem{wu2013online}
Y.~Wu, J.~Lim, and M.-H. Yang, ``Online object tracking: A benchmark,'' in {\em
  Proceedings of the IEEE conference on computer vision and pattern
  recognition}, pp.~2411--2418, 2013.

\bibitem{smeulders2013visual}
A.~W. Smeulders, D.~M. Chu, R.~Cucchiara, S.~Calderara, A.~Dehghan, and
  M.~Shah, ``Visual tracking: An experimental survey,'' {\em IEEE transactions
  on pattern analysis and machine intelligence}, vol.~36, no.~7,
  pp.~1442--1468, 2013.

\bibitem{breitenstein2009robust}
M.~D. Breitenstein, F.~Reichlin, B.~Leibe, E.~Koller-Meier, and L.~Van~Gool,
  ``Robust tracking-by-detection using a detector confidence particle filter,''
  in {\em 2009 IEEE 12th International Conference on Computer Vision},
  pp.~1515--1522, IEEE, 2009.

\bibitem{kalal2011tracking}
Z.~Kalal, K.~Mikolajczyk, and J.~Matas, ``Tracking-learning-detection,'' {\em
  IEEE transactions on pattern analysis and machine intelligence}, vol.~34,
  no.~7, pp.~1409--1422, 2011.

\bibitem{bewley2016simple}
A.~Bewley, Z.~Ge, L.~Ott, F.~Ramos, and B.~Upcroft, ``Simple online and
  realtime tracking,'' in {\em 2016 IEEE international conference on image
  processing (ICIP)}, pp.~3464--3468, IEEE, 2016.

\bibitem{bergmann2019tracking}
P.~Bergmann, T.~Meinhardt, and L.~Leal-Taixe, ``Tracking without bells and
  whistles,'' in {\em Proceedings of the IEEE/CVF International Conference on
  Computer Vision}, pp.~941--951, 2019.

\bibitem{wojke2017simple}
N.~Wojke, A.~Bewley, and D.~Paulus, ``Simple online and realtime tracking with
  a deep association metric,'' in {\em 2017 IEEE international conference on
  image processing (ICIP)}, pp.~3645--3649, IEEE, 2017.

\bibitem{Wojke2018deep}
N.~Wojke and A.~Bewley, ``Deep cosine metric learning for person
  re-identification,'' in {\em 2018 IEEE Winter Conference on Applications of
  Computer Vision (WACV)}, pp.~748--756, IEEE, 2018.

\bibitem{cao2022observation}
J.~Cao, X.~Weng, R.~Khirodkar, J.~Pang, and K.~Kitani, ``Observation-centric
  sort: Rethinking sort for robust multi-object tracking,'' {\em arXiv preprint
  arXiv:2203.14360}, 2022.

\bibitem{huang2021joint}
K.~Huang and Q.~Hao, ``Joint multi-object detection and tracking with
  camera-lidar fusion for autonomous driving,'' in {\em 2021 IEEE/RSJ
  International Conference on Intelligent Robots and Systems (IROS)},
  pp.~6983--6989, IEEE, 2021.

\bibitem{dewan2016motion}
A.~Dewan, T.~Caselitz, G.~D. Tipaldi, and W.~Burgard, ``Motion-based detection
  and tracking in 3d lidar scans,'' in {\em 2016 IEEE international conference
  on robotics and automation (ICRA)}, pp.~4508--4513, IEEE, 2016.

\bibitem{alvarez2019people}
C.~{\'A}lvarez-Aparicio, {\'A}.~M. Guerrero-Higueras, F.~J.
  Rodr{\'\i}guez-Lera, J.~Gin{\'e}s~Clavero, F.~Mart{\'\i}n~Rico, and
  V.~Matell{\'a}n, ``People detection and tracking using lidar sensors,'' {\em
  Robotics}, vol.~8, no.~3, p.~75, 2019.

\bibitem{osep2017combined}
A.~Osep, W.~Mehner, M.~Mathias, and B.~Leibe, ``Combined image-and world-space
  tracking in traffic scenes,'' in {\em 2017 IEEE International Conference on
  Robotics and Automation (ICRA)}, pp.~1988--1995, IEEE, 2017.

\bibitem{scheidegger2018mono}
S.~Scheidegger, J.~Benjaminsson, E.~Rosenberg, A.~Krishnan, and
  K.~Granstr{\"o}m, ``Mono-camera 3d multi-object tracking using deep learning
  detections and pmbm filtering,'' in {\em 2018 IEEE Intelligent Vehicles
  Symposium (IV)}, pp.~433--440, IEEE, 2018.

\bibitem{chen2011kalman}
S.~Chen, ``Kalman filter for robot vision: a survey,'' {\em IEEE Transactions
  on industrial electronics}, vol.~59, no.~11, pp.~4409--4420, 2011.

\bibitem{nguyen2004fast}
H.~T. Nguyen and A.~W. Smeulders, ``Fast occluded object tracking by a robust
  appearance filter,'' {\em IEEE transactions on pattern analysis and machine
  intelligence}, vol.~26, no.~8, pp.~1099--1104, 2004.

\bibitem{kalman1960new}
R.~E. Kalman, ``A new approach to linear filtering and prediction problems,''
  1960.

\bibitem{luiten2020track}
J.~Luiten, T.~Fischer, and B.~Leibe, ``Track to reconstruct and reconstruct to
  track,'' {\em IEEE Robotics and Automation Letters}, vol.~5, no.~2,
  pp.~1803--1810, 2020.

\bibitem{mao2022review3dod}
J.~Mao, S.~Shi, X.~Wang, and H.~Li, ``3d object detection for autonomous
  driving: A review and new outlooks,'' {\em arXiv preprint arXiv:2206.09474},
  2022.

\bibitem{beker2020monocular}
D.~Beker, H.~Kato, M.~A. Morariu, T.~Ando, T.~Matsuoka, W.~Kehl, and A.~Gaidon,
  ``Monocular differentiable rendering for self-supervised 3d object
  detection,'' in {\em European Conference on Computer Vision}, pp.~514--529,
  Springer, 2020.

\bibitem{ku2019monocular}
J.~Ku, A.~D. Pon, and S.~L. Waslander, ``Monocular 3d object detection
  leveraging accurate proposals and shape reconstruction,'' in {\em Proceedings
  of the IEEE/CVF conference on computer vision and pattern recognition},
  pp.~11867--11876, 2019.

\bibitem{he2019mono3d++}
T.~He and S.~Soatto, ``Mono3d++: Monocular 3d vehicle detection with two-scale
  3d hypotheses and task priors,'' in {\em Proceedings of the AAAI Conference
  on Artificial Intelligence}, vol.~33, pp.~8409--8416, 2019.

\bibitem{xiang2015voxel3dod}
Y.~Xiang, W.~Choi, Y.~Lin, and S.~Savarese, ``Data-driven 3d voxel patterns for
  object category recognition,'' in {\em Proceedings of the IEEE conference on
  computer vision and pattern recognition}, pp.~1903--1911, 2015.

\bibitem{ost2021neural}
J.~Ost, F.~Mannan, N.~Thuerey, J.~Knodt, and F.~Heide, ``Neural scene graphs
  for dynamic scenes,'' in {\em Proceedings of the IEEE/CVF Conference on
  Computer Vision and Pattern Recognition (CVPR)}, pp.~2856--2865, 2021.

\bibitem{zakharov2020autolabeling}
S.~Zakharov, W.~Kehl, A.~Bhargava, and A.~Gaidon, ``Autolabeling 3d objects
  with differentiable rendering of sdf shape priors,'' in {\em Proceedings of
  the IEEE/CVF Conference on Computer Vision and Pattern Recognition},
  pp.~12224--12233, 2020.

\bibitem{thies2019deferred}
J.~Thies, M.~Zollhöfer, and M.~Nießner, ``Deferred neural rendering,'' {\em
  ACM Transactions on Graphics}, vol.~38, p.~1–12, Jul 2019.

\bibitem{sitzmann2019deepvoxels}
V.~Sitzmann, J.~Thies, F.~Heide, M.~Nie{\ss}ner, G.~Wetzstein, and
  M.~Zollh{\"o}fer, ``Deepvoxels: Learning persistent 3d feature embeddings,''
  in {\em Proceedings of the IEEE/CVF Conference on Computer Vision and Pattern
  Recognition (CVPR)}, 2019.

\bibitem{yuan2021star}
W.~Yuan, Z.~Lv, T.~Schmidt, and S.~Lovegrove, ``Star: Self-supervised tracking
  and reconstruction of rigid objects in motion with neural rendering,'' in
  {\em Proceedings of the IEEE/CVF Conference on Computer Vision and Pattern
  Recognition}, pp.~13144--13152, 2021.

\bibitem{park2021nerfies}
K.~Park, U.~Sinha, J.~T. Barron, S.~Bouaziz, D.~B. Goldman, S.~M. Seitz, and
  R.~Martin-Brualla, ``Nerfies: Deformable neural radiance fields,'' {\em
  Proceedings of the IEEE International Conference on Computer Vision}, 2021.

\bibitem{kellnhofer2021neural}
P.~Kellnhofer, L.~C. Jebe, A.~Jones, R.~Spicer, K.~Pulli, and G.~Wetzstein,
  ``Neural lumigraph rendering,'' in {\em Proceedings of the IEEE/CVF
  Conference on Computer Vision and Pattern Recognition}, pp.~4287--4297, 2021.

\bibitem{chou2022gensdf}
G.~Chou, I.~Chugunov, and F.~Heide, ``Gensdf: Two-stage learning of
  generalizable signed distance functions,'' in {\em Proc. of Neural
  Information Processing Systems (NeurIPS)}, 2022.

\bibitem{xiang2021neutex}
F.~Xiang, Z.~Xu, M.~Hasan, Y.~Hold-Geoffroy, K.~Sunkavalli, and H.~Su,
  ``Neutex: Neural texture mapping for volumetric neural rendering,'' in {\em
  Proceedings of the IEEE/CVF Conference on Computer Vision and Pattern
  Recognition}, pp.~7119--7128, 2021.

\bibitem{koestler2022intrinsic}
L.~Koestler, D.~Grittner, M.~Moeller, D.~Cremers, and Z.~L{\"a}hner,
  ``Intrinsic neural fields: Learning functions on manifolds,'' {\em arXiv
  preprint arXiv:2203.07967}, vol.~2, 2022.

\bibitem{karras2019styleGAN}
T.~Karras, S.~Laine, and T.~Aila, ``A style-based generator architecture for
  generative adversarial networks,'' in {\em Proceedings of the IEEE/CVF
  conference on computer vision and pattern recognition}, pp.~4401--4410, 2019.

\bibitem{karras2020styleGAN2}
T.~Karras, S.~Laine, M.~Aittala, J.~Hellsten, J.~Lehtinen, and T.~Aila,
  ``Analyzing and improving the image quality of stylegan,'' in {\em
  Proceedings of the IEEE/CVF conference on computer vision and pattern
  recognition}, pp.~8110--8119, 2020.

\bibitem{ho2020denoising}
J.~Ho, A.~Jain, and P.~Abbeel, ``Denoising diffusion probabilistic models,''
  {\em Advances in neural information processing systems}, vol.~33,
  pp.~6840--6851, 2020.

\bibitem{nichol2021improved}
A.~Q. Nichol and P.~Dhariwal, ``Improved denoising diffusion probabilistic
  models,'' in {\em International Conference on Machine Learning},
  pp.~8162--8171, PMLR, 2021.

\bibitem{hao2021gancraft}
Z.~Hao, A.~Mallya, S.~Belongie, and M.-Y. Liu, ``Gancraft: Unsupervised 3d
  neural rendering of minecraft worlds,'' {\em arXiv preprint
  arXiv:2104.07659}, 2021.

\bibitem{wang2021nerfmm}
Z.~Wang, S.~Wu, W.~Xie, M.~Chen, and V.~A. Prisacariu, ``Nerf--: Neural
  radiance fields without known camera parameters,'' {\em arXiv preprint
  arXiv:2102.07064}, 2021.

\bibitem{yen2021inerf}
L.~Yen-Chen, P.~Florence, J.~T. Barron, A.~Rodriguez, P.~Isola, and T.-Y. Lin,
  ``inerf: Inverting neural radiance fields for pose estimation,'' in {\em 2021
  IEEE/RSJ International Conference on Intelligent Robots and Systems (IROS)},
  pp.~1323--1330, IEEE, 2021.

\bibitem{lin2021barf}
C.-H. Lin, W.-C. Ma, A.~Torralba, and S.~Lucey, ``Barf: Bundle-adjusting neural
  radiance fields,'' in {\em Proceedings of the IEEE/CVF International
  Conference on Computer Vision}, pp.~5741--5751, 2021.

\bibitem{nimier2021material}
M.~Nimier-David, Z.~Dong, W.~Jakob, and A.~Kaplanyan, ``{Material and Lighting
  Reconstruction for Complex Indoor Scenes with Texture-space Differentiable
  Rendering},'' in {\em Eurographics Symposium on Rendering - DL-only Track}
  (A.~Bousseau and M.~McGuire, eds.), The Eurographics Association, 2021.

\bibitem{guo2022nerfren}
Y.-C. Guo, D.~Kang, L.~Bao, Y.~He, and S.-H. Zhang, ``Nerfren: Neural radiance
  fields with reflections,'' in {\em Proceedings of the IEEE/CVF Conference on
  Computer Vision and Pattern Recognition}, pp.~18409--18418, 2022.

\bibitem{NimierDavidVicini2019Mitsuba2}
M.~Nimier-David, D.~Vicini, T.~Zeltner, and W.~Jakob, ``Mitsuba 2: A
  retargetable forward and inverse renderer,'' {\em Transactions on Graphics
  (Proceedings of SIGGRAPH Asia)}, vol.~38, Dec. 2019.

\bibitem{zhang2018perceptual}
R.~Zhang, P.~Isola, A.~A. Efros, E.~Shechtman, and O.~Wang, ``The unreasonable
  effectiveness of deep features as a perceptual metric,'' in {\em Proceedings
  of the IEEE conference on computer vision and pattern recognition (CVPR)},
  pp.~586--595, 2018.

\bibitem{simonyan2015deep}
K.~Simonyan and A.~Zisserman, ``Very deep convolutional networks for
  large-scale image recognition,'' 2015.

\bibitem{Kingma2015AdamAM}
D.~P. Kingma and J.~Ba, ``Adam: A method for stochastic optimization,'' {\em
  CoRR}, vol.~abs/1412.6980, 2015.

\bibitem{kuhn1955hungarian}
H.~W. Kuhn, ``The hungarian method for the assignment problem,'' {\em Naval
  research logistics quarterly}, vol.~2, no.~1-2, pp.~83--97, 1955.

\bibitem{zhou2019CenterPointVision}
X.~Zhou, D.~Wang, and P.~Kr{\"a}henb{\"u}hl, ``Objects as points,'' in {\em
  arXiv preprint arXiv:1904.07850}, 2019.

\bibitem{shen2021dmtet}
T.~Shen, J.~Gao, K.~Yin, M.-Y. Liu, and S.~Fidler, ``Deep marching tetrahedra:
  a hybrid representation for high-resolution 3d shape synthesis,'' {\em
  Advances in Neural Information Processing Systems}, vol.~34, pp.~6087--6101,
  2021.

\bibitem{laine2020modular}
S.~Laine, J.~Hellsten, T.~Karras, Y.~Seol, J.~Lehtinen, and T.~Aila, ``Modular
  primitives for high-performance differentiable rendering,'' {\em ACM
  Transactions on Graphics (TOG)}, vol.~39, no.~6, pp.~1--14, 2020.

\bibitem{shapenet2015}
A.~X. Chang, T.~Funkhouser, L.~Guibas, P.~Hanrahan, Q.~Huang, Z.~Li,
  S.~Savarese, M.~Savva, S.~Song, H.~Su, J.~Xiao, L.~Yi, and F.~Yu,
  ``{ShapeNet: An Information-Rich 3D Model Repository},'' Tech. Rep.
  arXiv:1512.03012 [cs.GR], Stanford University --- Princeton University ---
  Toyota Technological Institute at Chicago, 2015.

\bibitem{bernardin2006MOTA}
K.~Bernardin, A.~Elbs, and R.~Stiefelhagen, ``Multiple object tracking
  performance metrics and evaluation in a smart room environment,'' in {\em
  Sixth IEEE International Workshop on Visual Surveillance, in conjunction with
  ECCV}, vol.~90, Citeseer, 2006.

\end{thebibliography}





\end{document}